\documentclass[preprint,12pt]{elsarticle}
\usepackage{amssymb}
\usepackage[para,online,flushleft]{threeparttable}
\graphicspath{ {./figures/} }
\usepackage{hyperref}
\usepackage{float}
\usepackage{verbatim}
\restylefloat{figure}
\restylefloat{table}

\journal{Journal}

\usepackage{titlesec}
\titleclass{\subsubsubsection}{straight}[\subsection]
\newcounter{subsubsubsection}[subsubsection]
\renewcommand\thesubsubsubsection{\thesubsubsection.\arabic{subsubsubsection}}
\titleformat{\subsubsubsection}{\normalfont\normalsize\itshape}{\thesubsubsubsection.\space}{0em}{}
\titlespacing*{\subsubsubsection}{0pt}{2ex plus 1ex minus .2ex}{0.75ex plus .2ex}
\makeatletter
\def\toclevel@subsubsubsection{4}
\def\l@subsubsubsection{\@dottedtocline{4}{7em}{4em}}
\makeatother

\usepackage{amssymb}
\usepackage{pifont}
\newcommand{\cmark}{\ding{51}}%
\newcommand{\xmark}{\ding{55}}%

\usepackage[table]{xcolor}
\usepackage{graphicx}
\usepackage{amsmath,amssymb,amsfonts}
\usepackage{algorithmic}
\usepackage{graphicx}
\usepackage{textcomp}
\usepackage{booktabs}
\usepackage{xcolor}
\usepackage{array}
\usepackage{multirow}
\usepackage{url}
\usepackage{float}
\usepackage{pifont}

\floatstyle{plaintop}
\restylefloat{table}
\usepackage{caption}
\usepackage{subcaption}

\usepackage{booktabs}
\newcommand{\tabitem}{~~\llap{\textbullet}~~}

\newif\ifblackandwhite
\blackandwhitetrue
\usepackage{pdflscape}
\usepackage{colortbl}
\newcommand{\myrowcolour}{\rowcolor[gray]{0.925}}
\usepackage{booktabs}

\usepackage{adjustbox}
\usepackage{rotating}

\def\BibTeX{{\rm B\kern-.05em{\sc i\kern-.025em b}\kern-.08em
    T\kern-.1667em\lower.7ex\hbox{E}\kern-.125emX}}

\usepackage{geometry}
\begin{document}
\begin{frontmatter}

\begin{titlepage}
\begin{center}
\vspace*{0.5cm}

\textbf{Eyes on the Environment: AI-Driven Analysis for Fire and Smoke Classification, Segmentation, and Detection} \footnote{This material is based upon work supported by the National Aeronautics and Space Administration (NASA) under award number 80NSSC23K1393, and the National Science Foundation under Grant Numbers CNS-2232048, CNS 2120485, and CNS-2204445. 
}

\vspace{2cm}

Sayed Pedram Haeri Boroujeni$^{a*}$ (shaerib@g.clemson.edu)\\
Niloufar Mehrabi$^a$ (nmehrab@g.clemson.edu)\\
Fatemeh Afghah$^b$ (fafghah@clemson.edu)\\
Connor Peter McGrath$^c$ (mcgrat5@g.clemson.edu)\\
Danish Bhatkar$^a$ (dbhatka@g.clemson.edu)\\
Mithilesh Anil Biradar$^a$ (mbirada@clemson.edu)\\
Abolfazl Razi$^a$ (arazi@clemson.edu)\\

\hspace{10pt}

\begin{flushleft}
\small  
$^a$School of Computing, Clemson University, Clemson, SC 29632, USA\\[1mm]
$^b$Department of Electrical and Computer Engineering, Clemson University, Clemson, SC 29634, USA\\[1mm]

\vspace{2.5cm}

\end{flushleft}        
\end{center}
\end{titlepage}

\title{Eyes on the Environment: AI-Driven Analysis for Fire and Smoke Classification, Segmentation, and Detection}


\author{Sayed Pedram Haeri Boroujeni$^{a*}$, Niloufar Mehrabi$^{a}$, Fatemeh Afghah$^{b}$, Connor Peter McGrath$^{a}$, Danish Bhatkar$^{a}$, Mithilesh Anil Biradar$^{a}$, Abolfazl Razi$^{a}$}

\affiliation{organization={School of Computing},
            addressline={Clemson University}, 
            city={Clemson},
            postcode={29632}, 
            state={SC},
            country={USA}}

\affiliation{organization={Department of Electrical and Computer Engineering},
            addressline={Clemson University}, 
            city={Clemson},
            postcode={29634}, 
            state={SC},
            country={USA}}

\begin{abstract}
Fire and smoke phenomena pose a significant threat to the natural environment, ecosystems, and global economy, as well as human lives and wildlife. In this particular circumstance, there is a demand for more sophisticated and advanced technologies to implement an effective strategy for early detection, real-time monitoring, and minimizing the overall impacts of fires on ecological balance and public safety. Recently, the rapid advancement of Artificial Intelligence (AI) and Computer Vision (CV) frameworks has substantially revolutionized the momentum for developing efficient fire management systems. However, these systems extensively rely on the availability of adequate and high-quality fire and smoke data to create proficient Machine Learning (ML) methods for various tasks, such as detection and monitoring. Although fire and smoke datasets play a critical role in training, evaluating, and testing advanced Deep Learning (DL) models, a comprehensive review of the existing datasets is still unexplored. For this purpose, we provide an in-depth review to systematically analyze and evaluate fire and smoke datasets collected over the past 20 years. We investigate the characteristics of each dataset, including type, size, format, collection methods, and geographical diversities. We also review and highlight the unique features of each dataset, such as imaging modalities (RGB, thermal, infrared) and their applicability for different fire management tasks (classification, segmentation, detection). Furthermore, we summarize the strengths and weaknesses of each dataset and discuss their potential for advancing research and technology in fire management. Ultimately, we conduct extensive experimental analyses across different datasets using several state-of-the-art algorithms, such as ResNet-50, DeepLab-V3, and YoloV8. These experiments provide a comparative evaluation of the dataset's performance, emphasizing their strengths, limitations, and generalizability in diverse scenarios. 
\end{abstract}


\end{frontmatter}

\section{Introduction }
\label{sec:Introduction}

In recent years, fire and smoke have played a critical role as major global issues since their catastrophic effects on human life, ecosystems, and infrastructure are unavoidable \cite{boroujeni2024comprehensive}. Fires produce considerable levels of carbon dioxide and other greenhouse gases, which significantly impact climate change. Over time, this process can lead to more frequent and severe fires. Additionally, smoke from fires can harm human health, especially in vulnerable populations. Smoke can spread over long distances and affect air quality, allowing it to reach regions far from its source \cite{filonchyk2024changes}. It also seriously affects urban and industrial regions by causing property damage, loss of life, and economic hardship due to out-of-control fires. Fires not only cause immediate destruction but also lead to indirect consequences that amplify their overall impact. For instance, forest fires can alter water cycles, decrease agricultural yields, and accelerate desertification. Urban and industrial fires overwhelm emergency response systems, cause significant economic losses, and delay recovery efforts \cite{deng2025quantification}. In particular, smoke exacerbates disaster management challenges by reducing visibility, blocking evacuation routes, and hindering aerial firefighting operations. These complex challenges highlight the critical need for innovative technologies and methods to predict, monitor, and mitigate the effects of fire and smoke \cite{ribeiro2020forest}. 

\begin{figure}[H]
   \centering
   \centerline{\includegraphics[width=0.95\textwidth]{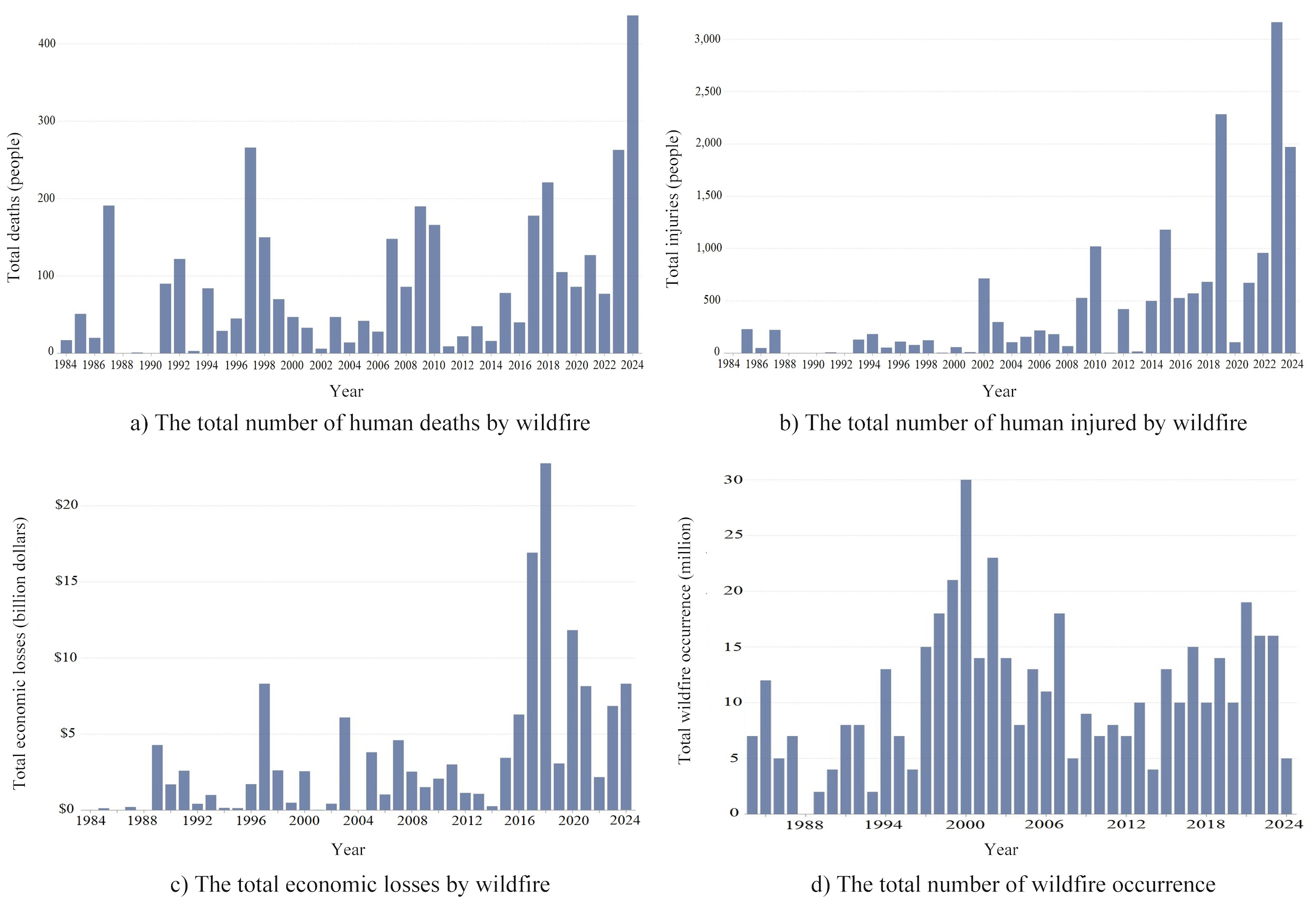}}
   \caption{The pattern of wildfire disasters between the years 1984 to 2024.}
   \label{fig: Fire Disaster Trend}
\end{figure}

In addition to providing information about the effects of fire and smoke phenomena, Figure \ref{fig: Fire Disaster Trend} shows some statistical analysis on wildfire trends during the period of 1984 to 2024. The figure highlights noteworthy statistics regarding the pattern of wildfire occurrences and its impacts on both humans and the economy \cite{sannigrahi2022examining}. According to Figure \ref{fig: Fire Disaster Trend}, there has been a substantial surge in the occurrence of wildfire hazards, with a 3-fold increase over the period of 1984 to 2024. Meanwhile, the total number of people affected and injured by this catastrophic event is marginally above the annual average between 1984 and 2024. Furthermore, despite a decline in economic losses during the last 10 years, the overall amount is still higher than the annual average from 1984 to 2024. These statistics indicate that wildfire disasters are becoming more frequent, and despite a decline in economic impacts, their effects are nonetheless profound. Overall, these statistics underline the urgent necessity for more efficient fire management to mitigate the economic damages and long-term consequences impacted by this natural disaster. In response to this trend, researchers and policymakers from various fields of science have been noticeably attracted to focus on developing new strategies for understanding, managing, and mitigating the consequences of fire disasters \cite{wang2023model, ozek2023examining}.

\vspace{10mm}

Traditional methods, such as ground cameras and manual reporting, have been used to monitor and assess fire and smoke. However, these approaches are significantly constrained in coverage, accuracy, and speed, mainly for large or remote areas. Ground cameras can miss important details and struggle to provide real-time data across vast regions. These obstacles complicate the effective control of flames in dynamic and broad ecosystems. Advanced methodologies, including Deep Learning (DL) \cite{boroujeni2024ic}, Computer Vision (CV) \cite{wang2024flame}, and drone-based surveillance \cite{mehrabi2024adaptive}, enable the processing of large datasets and the analysis of complex patterns and deliver real-time insights, which make them crucial for enhancing fire detection, monitoring, and response. These technologies enable faster, more accurate assessments of fire dynamics, smoke dispersion, and risk prediction, which are critical for effective disaster management and mitigation. Comprehensive databases are key to understanding the full impact of wildfires and supporting further research.

Datasets, which must be large, diverse, and well-annotated, are critical for training and evaluating models that advance real-time fire detection, smoke dispersion analysis, and fire behavior prediction with unprecedented accuracy and efficiency. Researchers have used these data sets to develop systems to support applications such as resource allocation in emergency situations, environmental monitoring, and decision-making in disaster management \cite{zhou2024characterizing}. Datasets encompassing diverse environmental conditions, fire intensities, and smoke characteristics enable researchers to understand fire dynamics better and develop scalable and generalizable solutions applicable to natural and human-made fire scenarios. Researchers face significant challenges in collecting and curating datasets for fire and smoke, as these datasets often differ across several critical dimensions. Producing datasets that accurately represent real-world scenarios presents a significant challenge due to environmental factors such as lighting, weather, and topography variations. Moreover, gathering fire and smoke data frequently necessitates working in hazardous and remote environments, which leads to logistical and safety constraints \cite{heibati2025wildfire}. 

Datasets can be categorized in various ways based on the type of information they contain, such as RGB, thermal, multispectral, hyperspectral, and infrared images \cite{boroujeni2024comprehensive,Shamsoshoara2023}. Consequently, specialized sensors and processing methodologies are necessary to capture fire and smoke under diverse situations. For example, Infrared (IR) cameras are commonly employed to identify heat signatures from flames, especially in low visibility situations like nighttime or through smoke. Thermal cameras assist in identifying temperature fluctuations in fire-impacted regions, whilst multispectral or hyperspectral sensors may discern various types of smoke and particle matter by collecting data across numerous light wavelengths. Moreover, LiDAR (Light Detection and Ranging) sensors provide three-dimensional mapping of the topography and fire propagation, supplying essential information for wildfire management. In addition, radar-based systems, such as Doppler radar, can track smoke movement and dispersion patterns which support real-time monitoring of fire behavior \cite{fulton2024use}.

The quality and resolution of the data directly affect the accuracy and dependability of models trained on these datasets. While higher-resolution datasets provide greater detail, they also require more computational resources. The dataset's size—in terms of the number of photos or video frames—exhibits considerable variability as some provide a broad coverage while others may be restricted to particular situations or areas. The variation exists even with respect to the image-based data set versus the video-based data set because a video will have essential temporal information in the study of the progress of a fire or the dynamics of smoke flow but demands higher storage capacity and more processing power \cite{alzorgan2023actuator}. Labeling procedures inside datasets is another notable distinction. Some datasets provide simple binary annotations, such as “fire” versus “no fire,” while others offer more detailed labels, such as “fire,” “smoke,” or “no fire/smoke” \cite{boroujeni2024comprehensive}. In some cases, datasets include additional categories such as “fire,” “smoke,” “none,” and “both” to represent different scenarios. Moreover, datasets may differ in annotation quality and diversity, with some offering thorough segmentation masks or bounding boxes while others depend on less accurate annotations. The discrepancies in data types, quality, and annotation methods significantly lead to a massive gap in practicality and applicability for various tasks, including fire detection, smoke segmentation, and behavior prediction \cite{aragoneses2025multi}. Furthermore, the need for established benchmarks and assessment criteria complicates model comparisons across diverse datasets, impeding progress in research and the creation of generally applicable solutions. 

These issues must be addressed by testing the properties and usability of each dataset for a broad spectrum of applications. This study categorizes and analyzes datasets to help researchers choose the best relevant datasets and discover regions requiring further datasets.

The major contributions of this review paper can be summarized as follows:

\begin{itemize}

    \item Provides an in-depth and systematic analysis of fire and smoke datasets collected over the past 20 years. We highlight the key characteristics of each dataset, such as type, size, format, perspective, collection methods, and geographical diversity.
   
    \item Explores the unique features of each dataset, including imaging modalities (e.g., RGB, thermal, infrared) and their successful applications in various fire management tasks, such as classification, segmentation, and detection.

    \item Summarizes the strengths and weaknesses of each dataset and assesses their potential to advance research and technology in fire management. We also provide valuable insights into the gaps and opportunities for developing more comprehensive datasets to address current challenges in the field.

    \item Performs extensive experimental evaluations on various datasets using well-known state-of-the-art algorithms, including ResNet-50, DeepLab-V3, and YOLOv8. Additionally, we discuss comparative performance evaluations across datasets, highlighting their capabilities, deficiencies, and generalizability in diverse scenarios.
    
    \item Outlines open problems and future directions in fire dataset research to facilitate the development of advanced AI and CV-based fire management systems. These directions include leveraging domain adaptation techniques, improving dataset diversity and quality, integrating spatiotemporal data, and designing benchmarks to evaluate model generalizability.
    
\end{itemize}

\begin{figure}[H]
   \centering
   \centerline{\includegraphics[width=\textwidth]{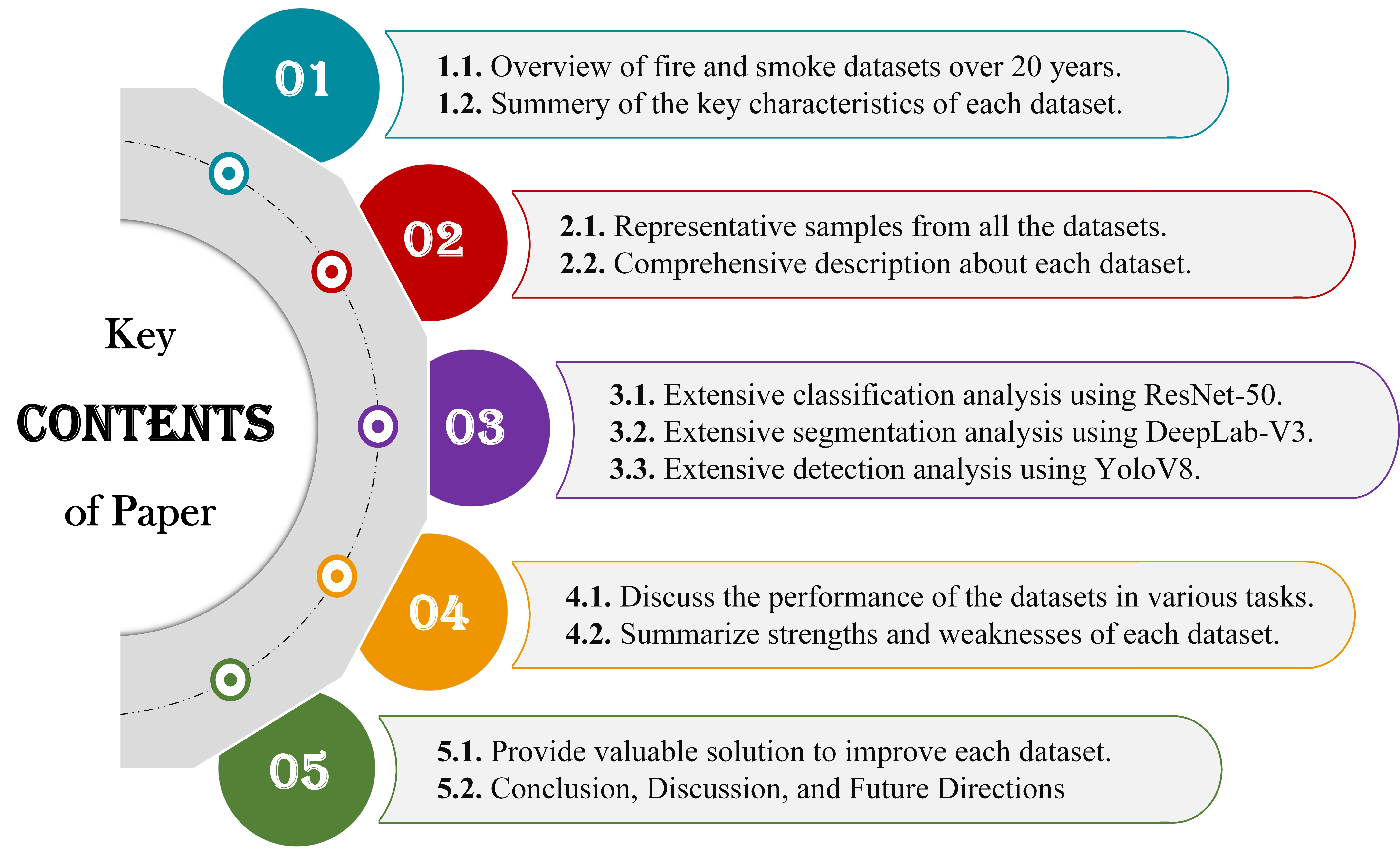}}
   \caption{The key contents covered by this review paper.}
   \label{fig: content}
\end{figure}

The general structure of the paper is presented in Figure \ref{fig: content}. Section \ref{sec: Wildfire Dataset} reviews the existing fire and smoke datasets over the past 20 years, followed by the strengths and weaknesses of each dataset. The experimental analysis and discussion are presented in Section \ref{sec: result}. Lastly, Section \ref{sec: Conclusion} provides conclusions and future directions.

\section{Wildfire Datasets}
\label{sec: Wildfire Dataset}

In the realm of wildfire research, the availability and quality of datasets play a pivotal role in advancing our understanding of fire dynamics, risk assessment, and mitigation strategies. This section provides a comprehensive review of existing wildfire datasets, highlighting their background, characteristics, applications, and strengths and limitations. Additionally, these datasets have been collected through different methods and sources, resulting in variations in accuracy and resolution. These variations in accuracy and resolution can affect the reliability and applicability of the datasets for different research purposes. Therefore, through this section, we examine each dataset's key attributes, including the number of samples, labels, data types, and the associated methodologies used in data collection. By delving into the details of these datasets, we aim to provide a comprehensive overview that will guide researchers in selecting the most relevant and robust datasets for their specific analyses and contribute to the ongoing efforts to enhance wildfire modeling and management practices.

This work compiles all existing wildfire datasets, encompassing those utilized in experimental studies related to wildfire classification, segmentation, and detection in various research papers. This collection aims to contribute to the comprehensive landscape of wildfire research, facilitating further advancements in this critical field.

Table \ref{Table:Dataset Review} describes a comprehensive overview of the most significant fire and smoke datasets, highlighting their release year, full name, category, and applications in classification, segmentation, and detection tasks. These datasets are categorized as Original, Aggregated, or Mixed, emphasizing the source and uniqueness of the data. Original datasets are exclusively collected by the authors, Aggregated datasets are entirely extracted from existing sources, and Mixed datasets integrate data from both categories.

\newcommand{\cxmark}{\ding{51}\hspace{-1.75mm}\ding{55}}
\begin{table}[H]
\caption{Previous fire and smoke datasets overview.}
\vspace{2mm}

\label{Table:Dataset Review}
\begin{center}
\resizebox{\textwidth}{!}{
\setlength{\tabcolsep}{4pt}
\begin{tabular}{lllllcccc}
\toprule
\multicolumn{2}{c}{} & \multicolumn{1}{c}{} & \multicolumn{2}{c}{\textbf{}} &\multicolumn{3}{c}{\textbf{Application}} \\ [1.5mm]

\cmidrule(l){6-8}
\multirow{1}{*}{\textbf{Ref}}   & \textbf{Year}    & \textbf{Full Name}   & \textbf{Arbitrary}              & \textbf{Dataset Category}  
                                      & \textbf{Classification}         & \textbf{Segmentation}          & \textbf{Detection} \\ [1.5mm]

\midrule 

\cite{hopkins2024flame3datasetunleashing}      &2024      &Fire modeLing Aerial Multi-spectral imagE3     &FLAME3      &Original Dataset$^*$     &\cmark     &\xmark   &\cmark \\[3mm]

\cite{wang2024flame}      &2024      &Flame Standard Diffuser Dataset    &FLAME-SD      &Generated Dataset$^*$      &\cmark     &\xmark   &\cmark \\[3mm]

\cite{wu2023dataset}      &2023      &Dataset for Fire and Smoke Detection     &DFS      &Aggregated Dataset$^*$     &\cmark     &\xmark   &\cmark \\[3mm]

\cite{ribeiro2023burned}    &2023    &Burned Area UAV Dataset     &BA-UAV       &Original Dataset     &\xmark     &\cmark    &\xmark      \\[3mm]

\cite{wang2023efficient}   &2023     &Fire Detection Dataset      &FireDetn     &Aggregated Dataset     &\xmark      &\xmark     &\cmark          \\[3mm]

\cite{hu2023firefly}   &2023     &Synthetic Dataset for Ember Detection      &FireFly     &Generated Dataset     &\xmark      &\xmark     &\cmark          \\[3mm]

\cite{chen2022wildland}   &2022      &Fire modeLing Aerial Multi-spectral imagE2      &FLAME2     &Original Dataset     &\cmark     &\xmark     &\xmark       \\[3mm]

\cite{ribeiro2022towards}    &2022    &Forest Fire Mapping and Spread Forecast     &FireFront     &Original Dataset     &\cmark     &\xmark    &\xmark         \\[3mm]

\cite{khan2022deepfire}     &2022    &Deep Transfer Learning Benchmark    &DeepFire    &Aggregated Dataset     &\cmark     &\xmark    &\xmark      \\[3mm]

\cite{Peach}      &2021    &Paddle Fire Dataset    &Paddle     &Aggregated Dataset     &\cmark     &\xmark    &\xmark     \\[3mm]

\cite{datacluster} &2021      &Data Cluster Labs' Fire and Smoke Dataset   &DataCluster      &Original Dataset     &\cmark     &\xmark    &\cmark          \\[3mm]

\cite{dincer}    &2021   &Wildfire Detection Image Data  &Kaggle            &Aggregated Dataset       &\cmark     &\xmark    &\xmark          \\[3mm]

\cite{li2020efficient}   &2020   &Fire Detection Dataset        &FD-Dataset      &Aggregated Dataset           &\xmark      &\xmark     &\cmark           \\[3mm]

\cite{de2022automatic}   &2020    &Fire and Smoke Detection Dataset    &D-Fire         &Mixed Dataset$^*$         &\cmark     &\xmark    &\cmark         \\[3mm]

\cite{shamsoshoara2021aerial}  &2020      &Fire modeLing Aerial Multi-spectral imagE1    &FLAME1  &Original Dataset    &\cmark      & \cmark    &\xmark           \\[3mm]

\cite{khanali}     &2020    &Dataset for Forest Fire Detection    &FF-Det     &Aggregated Dataset      &\cmark     &\xmark    &\xmark          \\[3mm]

\cite{Moses}      &2019     & Real-time Fire Detection Dataset       &FireNet         &Aggregated Dataset     &\xmark     &\xmark        &\cmark          \\[3mm]
     
\cite{kyrkou2019deep}    &2019  &Aerial Image Database for Emergency Response      & AIDER     &Aggregated Dataset      &\cmark     &\xmark    &\xmark          \\[3mm]

\cite{ForestryImages}   &2018  &Forestry Fire Image Dataset     &ForestryImage         &Mixed Dataset     &\cmark      & \xmark    &\xmark           \\[3mm]

\cite{Saied}   &2018   &Outdoor and Non-FIRE Image Dataset    &FIRE       &Aggregated Dataset     &\cmark     &\xmark    &\xmark          \\[3mm]

\cite{sharma}   &2017    &Centre for Artificial Intelligence Research     &CAIR          &Aggregated Dataset     &\cmark     &\xmark    &\xmark          \\[3mm]

\cite{braovic2017cogent}    &2017  &FESB Mediterranean Landscape Image Dataset     &FESB MLID          &Original Dataset       &\cmark     &\cmark   &\xmark         \\[3mm]

\cite{cazzolato2017fismo} &2017   &Fire and Smoke Images and Videos Dataset   &FiSmo        &Aggregated Dataset     &\cmark     & \xmark   & \xmark         \\[3mm]

\cite{toulouse2017computer} &2017    &Corsican Fire Image Database         &Corsican        &Original Dataset     &\xmark      &\cmark     &\xmark           \\[3mm]

\cite{grammalidis2017firesense} &2017    &Database of Videos for Flame/Smoke Detection     &FireSense       &Original Dataset       &\cmark      &\xmark     &\xmark           \\[3mm] 

\cite{chino2015bowfire}   &2015    &Best of Both Worlds Fire detection       &BoWFire        &Mixed Dataset     &\cmark     &\cmark    &\xmark          \\[3mm]

\cite{csvt2015_fire}    &2014     &Mivia's Fire Detection Dataset      &MIVIA           &Mixed Dataset     &\cmark       &\xmark      &\xmark            \\[3mm]

\cite{VisiFire}   &2010     &Computer Vision Based Fire Detection Dataset     &VisiFire        &Mixed Dataset     &\cmark     &\xmark    &\xmark          \\[3mm]

\cite{Cetin}    &2006    &Fire Video Clips Dataset     &FireClips         &Mixed Dataset     &\cmark     &\xmark    &\xmark          \\[3mm]

\bottomrule
\end{tabular}}
\end{center}
\footnotesize{\vspace{0.5mm}
$^*$ Dataset Category ``\textbf{Original}'' denotes that the data were collected and generated exclusively by the authors, emphasizing the dataset's originality. ``\textbf{Generated}'' signifies that the data are artificially created. ``\textbf{Mixed}'' indicates that some data was originally collected by the authors, while others were sourced from existing datasets or the internet. The term ``\textbf{Aggregated}'' refers to the corresponding dataset where the entire data is selected from various existing sources, such as other datasets or the internet, without implying any original content from the authors.}
\end{table}

Table \ref{Table: Dataset} provides a comprehensive summary and comparative analysis of the significant fire and smoke datasets. Key attributes such as dataset size, labeling details, collection methods, and accessibility are summarized, as well as highlighting their geographical diversities and acquisition techniques. These datasets are valuable resources for advancing fire management, smoke analysis, and environmental monitoring technologies. Additionally, Table \ref{Table: DatasetLink} specifies complete download links for accessing the full dataset contents and information.

\begin{table*}[htbp]
\caption{Summary and comparative analysis of the significant wildland fire and smoke datasets.}
\label{Table: Dataset}
\begin{center}
\resizebox{\textwidth}{!}{
\setlength{\tabcolsep}{4pt}
\begin{tabular}{llllllllllc}
\toprule

\multirow{1}{*}{\textbf{No.}}      & \textbf{Year}   & \textbf{Dataset}   & \textbf{Dataset}            & 
     \textbf{Dataset}               &   \textbf{Dataset}  
                                      & \textbf{Dataset}         & \textbf{Labeling}            & \textbf{Location}    & \textbf{Collection}    & \textbf{Access}\\

                 \textbf{}        & \textbf{}        & \textbf{Name}         & \textbf{Type}          
                       &\textbf{Format}        & \textbf{Perspective} 
                                      & \textbf{Size} &\textbf{Type}   & \textbf{Area}        & \textbf{Method}      & \textbf{Link}\\

\midrule

\multirow{1}{*}{1} &2024  &FLAME3    &Image,  &.JPG  &Aerial   & 13.997    &Fire, Smoke,    &OR, USA  &UAV 
&\textcolor{blue}{\href{https://www.kaggle.com/datasets/brycehopkins/flame-3-computer-vision-subset-sycan-marsh}{Download}} \\[0mm]

&  &    &   &.TIFF  &  &Images   &No Fire/Smoke    &FL, USA  &Footage    &  \\[1mm]

\myrowcolour
\multirow{1}{*}{2} &2024  &FLAME-SD    &Image,  &.PNG  &Aerial   & 10,000    &Fire, Smoke,    &Forest  &Synthetic &\textcolor{blue}{\href{https://github.com/AIS-Clemson/FLAME_SD}{Download}}   \\[0mm]
\myrowcolour
&  &    &   &  &  &Images   &No Fire/Smoke    &Wildfire  &Dataset    &  \\[1mm]

\multirow{1}{*}{3} &2023  &DFS    &Image,  &.PNG  &Terrestrial   & 9,462    &Fire, Smoke,    &Rural,  &Datasets, &\textcolor{blue}{\href{https://github.com/siyuanwu/DFS-FIRE-SMOKE-Dataset?tab=readme-ov-file}{Download}}   \\[0mm]

&  &    &Video   &.MP4  &  &Images   &Other    &Urban  &Internet    &  \\[1mm]

\myrowcolour
\multirow{1}{*}{4} &2023  &BA-UAV   &Image,   &.JPG     &Aerial   &22,500    &Burned,    &Portugal  &UAV  &\textcolor{blue}{\href{https://github.com/CIIC-C-T-Polytechnic-of-Leiria/ESS_Data_Lab}{Download}}  \\[0mm]
\myrowcolour
&  &    &Video  &.MP4  &  &Frames    &Unburned    &Wildfire  &Footage &  \\[1mm]

\multirow{1}{*}{5} &2023  &FireDetn    &Image   &.JPG  &Mixed   &4,674    &Fire,    &Rural  &Search &\textcolor{blue}{\href{https://github.com/SuperXxts/FireDetn}{Download}}  \\[0mm]

&  &    &  &   &  &Images    &No Fire    &Wildfire  &Engines &  \\[1mm]

\myrowcolour
\multirow{1}{*}{6} &2023  &FireFly    &Image,  &.JPG   &Aerial   &19,273    &Fire, Smoke,    &Forest  &Synthetic &\textcolor{blue}{\href{https://github.com/ERGOWHO/Firefly2.0}{Download}}   \\[0mm]
\myrowcolour
&  &    &   &  &  &Frames    &No Fire/Smoke    &Wildfire  &Dataset  &  \\[1mm]

\multirow{1}{*}{7} &2022  &FLAME2    &Image,  &.JPG   &Aerial   &53,451    &Fire, Smoke,    &AZ, USA  &Drone &\textcolor{blue}{\href{https://ieee-dataport.org/open-access/flame-2-fire-detection-and-modeling-aerial-multi-spectral-image-dataset}{Download}}   \\[0mm]

&  &    &Video   &.MP4  &  &Frames    &No Fire/Smoke    &Wildfire  &Footage  &  \\[1mm]

\myrowcolour
\multirow{1}{*}{8} &2022  &FireFront   &Video  &.MP4   &Aerial   &5    &Fire,    & Rural &UAV &\textcolor{blue}{\href{http://firefront.pt/}{Download}}  \\[0mm]
\myrowcolour
&  &    &  &  &  &Videos    &No Fire    &Wildfire  &Footage  &  \\[1mm]

\multirow{1}{*}{9} &2022  &DeepFire     &Image &.JPG    &Mixed    &1,900    &Fire,    &Rural  &Search &\textcolor{blue}{\href{https://www.kaggle.com/datasets/alik05/forest-fire-dataset}{Download}}  \\[0mm]

&  &    &    &   &  &Images    &No Fire    &Wildfire  &Engines  &  \\[1mm]

\myrowcolour
\multirow{1}{*}{10} &2021  &Paddle    &Image  &.PNG   &Mixed    &3,701    &Fire,    &Rural,  &Search &\textcolor{blue}{\href{https://aistudio.baidu.com/datasetdetail/107770}{Download}}   \\[0mm]
\myrowcolour
&  &    &  &  &  &Images    &No Fire    &Urban  &Engines  &  \\[1mm]

\multirow{1}{*}{11} &2021  &DataCluster    &Image   & .JPG  &Terrestrial   &7,000    &Fire,    &Rural,  &Cellphone &\textcolor{blue}{\href{https://www.kaggle.com/datasets/dataclusterlabs/fire-and-smoke-dataset}{Download}}  \\[0mm]

&  &    &  &   &  &Images    &No Fire    &Urban  &Cameras  &  \\[1mm]

\myrowcolour
\multirow{1}{*}{12} &2021  &Kaggle    &Image   &.JPG     &Mixed    &1,900    &Fire,    &Rural  &Search &\textcolor{blue}{\href{https://www.kaggle.com/datasets/brsdincer/wildfire-detection-image-data}{Download}}  \\[0mm]
\myrowcolour
&  &    &    &     &    &Images    &No Fire    &Wildfire  &Engines  &  \\[1mm]

\multirow{1}{*}{13} &2020  &FD-Dataset    &Image,    &.JPG   &Terrestrial   &50,000    &Fire,    &Rural,  &Search   &\textcolor{blue}{\href{http://www.nnmtl.cn/EFDNet/}{Download}}  \\[0mm]

&  &    &Video   &.MP4  &  &Frames    &No Fire    &Urban  &Engines  &  \\[1mm]

\myrowcolour
\multirow{1}{*}{14} &2020  &D-Fire    &Image     &.JPG   &Mixed    &21,527    &Fire, Smoke,    &Brazil  &Camera, &\textcolor{blue}{\href{https://github.com/gaiasd/DFireDataset}{Download}}  \\[0mm]
\myrowcolour
&  &    &Video  &.MP4  &  &Frames    &None, Both    &Wildfire  &Internet  &  \\[1mm]

\multirow{1}{*}{15} &2020  &FLAME1    &Image,   &.JPG  &Aerial   &47,992    &Fire,    &AZ, USA  &UAV &\textcolor{blue}{\href{https://ieee-dataport.org/open-access/flame-dataset-aerial-imagery-pile-burn-detection-using-drones-uavs}{Download}}    \\[0mm]

&  &    &Video  &.MP4  &  &Frames    &No Fire    &Wildfire  &Footage  &  \\[1mm]

\myrowcolour
\multirow{1}{*}{16} &2020  &FF-Det   &Image   &.JPG  &Aerial   &1,900    &Fire,    &Rural   &Search  &\textcolor{blue}{\href{https://data.mendeley.com/datasets/gjmr63rz2r/1}{Download}}  \\[0mm]
\myrowcolour
&  &    &   & &  &Images    &No Fire    &Wildfire  &Engines  &  \\[1mm]

\multirow{1}{*}{17} &2019  &FireNet    &Image    &.JPG  &Terrestrial   &502    &Fire,    &Rural,  &Search   &\textcolor{blue}{\href{https://github.com/OlafenwaMoses/FireNet}{Download}}   \\[0mm]

&  &    &  &  &  &Images    &No Fire    &Urban  &Engines  &  \\[1mm]

\myrowcolour
\multirow{1}{*}{18} &2019  &AIDER   &Image   &.JPG   &Aerial   &1,000    &Fire,    &Rural,  &Search &\textcolor{blue}{\href{https://github.com/ckyrkou/AIDER}{Download}}  \\[0mm]
\myrowcolour
&  &    &  &  &  &Images    &No Fire    &Urban  &Engines  &  \\[1mm]

\multirow{1}{*}{19} &2018  &ForestryImage    &Image   &.JPG  &Mixed    &317,921    &Fire,    &Rural  &Camera, &\textcolor{blue}{\href{https://www.forestryimages.org/browse/subthumb.cfm?sub=740}{Download}}  \\[0mm]

&  &    &  &  &  &Images    &No Fire    &Wildfire  &Internet  &  \\[1mm]

\myrowcolour
\multirow{1}{*}{20} &2018  &FIRE  &Image   &.PNG  &Terrestrial   &999    &Fire,    &Rural  &Search  &\textcolor{blue}{\href{https://www.kaggle.com/datasets/phylake1337/fire-dataset}{Download}}  \\[1mm]
\myrowcolour
&  &    &   &   &  &Images    &No Fire    &Wildfire  &Engines  &  \\[1mm]

\multirow{1}{*}{21} &2017  &CAIR   &Image    &.JPG &Terrestrial   &651    &Fire,    &Mainly  &Search &\textcolor{blue}{\href{https://github.com/cair/Fire-Detection-Image-Dataset}{Download}}  \\[0mm]

&  &    &    &   &  &Images    &No Fire    &Urban  &Engines  &  \\[1mm]

\myrowcolour
\multirow{1}{*}{22} &2017  &FESB MLID   &Image,   &.JPEG  &Terrestrial   &400    &Smoke,    &Mainly  &Camera,     &\textcolor{blue}{\href{http://wildfire.fesb.hr}{Download}}  \\[0mm]
\myrowcolour
&  &    &Video  &.MP4  &  &Images    &No Smoke    &Rural  &Sensors  &  \\[1mm]

\multirow{1}{*}{23} &2017  &FiSmo   &Image,   &.JPG  &Terrestrial   &9,448    &Fire, Smoke    &Rural,  &Search    &\textcolor{blue}{\href{https://github.com/mtcazzolato/dsw2017}{Download}}  \\[0mm]

&  &    &Video   &.MP4    &  &Images    &No Fire/Smoke    &Urban  &Engines  &  \\[1mm]

\myrowcolour
\multirow{1}{*}{24} &2017  &Corsican    &Image   &.JPG  &Terrestrial   &500    &Fire,    &Rural  &Camera, &\textcolor{blue}{\href{https://cfdb.univ-corse.fr/index.php}{Download}}  \\[0mm]
\myrowcolour
&  &    &Video  &.MP4  &  &Images    &No Fire   &Wildfire  &Sensors  &  \\[1mm]

\multirow{1}{*}{25} &2017  &FireSense    &Video  &.AVI   &Terrestrial    &49    &Fire, Smoke    &Rural   &Camera, &\textcolor{blue}{\href{https://zenodo.org/records/836749}{Download}}  \\[0mm]

&  &    &    &   &  &Videos    &No Fire/Smoke    &Urban  &Sensors  &  \\[1mm]

\myrowcolour
\multirow{1}{*}{26} &2015  &BoWFire    &Image  &.JPG   &Terrestrial   &466    &Fire,    &Rural   &Camera &\textcolor{blue}{\href{https://bitbucket.org/gbdi/bowfire-dataset/downloads/}{Download}}  \\[0mm]
\myrowcolour
&  &    &   &  &  &Images    &No Fire    &Urban  &Internet  &  \\[1mm]

\multirow{1}{*}{27} &2014  &MIVIA    &Image,  &.JPG    &Mixed    &62,690    &Fire,    &Rural  &Camera, &\textcolor{blue}{\href{https://mivia.unisa.it/datasets/video-analysis-datasets/fire-detection-dataset/}{Download}}   \\[0mm]

&  &    &Video   &.MP4  &  &Frames    &No Fire    &Urban  &Internet  &  \\[1mm]

\myrowcolour
\multirow{1}{*}{28} &2010  &VisiFire   &Video  &.MP4   &Terrestrial   &12    &Fire,    &Rural    &Camera, &\textcolor{blue}{\href{http://signal.ee.bilkent.edu.tr/VisiFire/}{Download}}   \\[0mm]
\myrowcolour
&  &    &    &   &  &Videos    &No Fire    &Urban  &Sensors  &  \\[1mm]

\multirow{1}{*}{29} &2006  &FireClips    &Video   &.AVI    &Terrestrial   &13    &Fire,    &Rural  &Camera, &\textcolor{blue}{\href{http://signal.ee.bilkent.edu.tr/VisiFire/Demo/FireClips/}{Download}}   \\[0mm]

&  &    &       &   &  &Videos    &No Fire    &Wildfire  &Sensors  &  \\[0mm]

\bottomrule
\end{tabular}}
\end{center}
\footnotesize{\vspace{0.5mm}

}
\end{table*}

\begin{table*}[t]
\caption{Complete access links for downloading the full dataset contents.}
\label{Table: DatasetLink}
\begin{center}
\resizebox{\textwidth}{!}{
\setlength{\tabcolsep}{4pt}
\begin{tabular}{llllllllc}
\toprule

\multirow{1}{*}{\textbf{No.}}     & \textbf{Dataset}         & \textbf{Download Link}  \\

\midrule 

1     &FLAME3         &\href{https://www.kaggle.com/datasets/brycehopkins/flame-3-computer-vision-subset-sycan-marsh}{https://www.kaggle.com/datasets/brycehopkins/flame-3-computer-vision-subset-sycan-marsh} \\[2mm]

\myrowcolour
2     &FLAME-SD      &\href{https://github.com/AIS-Clemson/FLAME_SD}{https://github.com/AIS-Clemson/FLAME-SD}    \\[2mm]

3     &DFS      &\href{https://github.com/siyuanwu/DFS-FIRE-SMOKE-Dataset?tab=readme-ov-file}{https://github.com/siyuanwu/DFS-FIRE-SMOKE-Dataset?tab=readme-ov-file}     \\[2mm]

\myrowcolour
4     &BA-UAV      &\href{https://github.com/ipleiria-ciic/ees-datalab}{https://github.com/ipleiria-ciic/ees-datalab}     \\[2mm]

5     &FireDetn      &\href{https://github.com/SuperXxts/FireDetn}{https://github.com/SuperXxts/FireDetn}     \\[2mm]

\myrowcolour
6     &FireFly      &\href{https://github.com/ERGOWHO/Firefly2.0}{https://github.com/SuperXxts/FireDetn}     \\[2mm]

7     &FLAME2      &\href{https://ieee-dataport.org/open-access/flame-2-fire-detection-and-modeling-aerial-multi-spectral-image-dataset}{https://ieee-dataport.org/open-access/flame-2-fire-detection-and-modeling-aerial-multi-spectral-image-dataset}     \\[2mm]

\myrowcolour
8     &FireFront      &\href{http://firefront.pt/}{http://firefront.pt/}     \\[2mm]

9     &DeepFire      &\href{https://www.kaggle.com/datasets/alik05/forest-fire-dataset}{https://www.kaggle.com/datasets/alik05/forest-fire-dataset}     \\[2mm]

\myrowcolour
10     &Paddle      &\href{https://aistudio.baidu.com/datasetdetail/107770}{https://aistudio.baidu.com/datasetdetail/107770}     \\[2mm]

11    &DataCluster      &\href{https://www.kaggle.com/datasets/dataclusterlabs/fire-and-smoke-dataset}{https://www.kaggle.com/datasets/dataclusterlabs/fire-and-smoke-dataset}     \\[2mm]

\myrowcolour
12    &Kaggle      &\href{https://www.kaggle.com/datasets/brsdincer/wildfire-detection-image-data}{https://www.kaggle.com/datasets/brsdincer/wildfire-detection-image-data}    \\[2mm]

13    &FD-Dataset      &\href{http://www.nnmtl.cn/EFDNet/}{http://www.nnmtl.cn/EFDNet/}    \\[2mm]

\myrowcolour
14    &D-Fire      &\href{https://github.com/gaiasd/DFireDataset}{https://github.com/gaiasd/DFireDataset}    \\[2mm]

15    &FLAME1      &\href{https://ieee-dataport.org/open-access/flame-dataset-aerial-imagery-pile-burn-detection-using-drones-uavs}{https://ieee-dataport.org/open-access/flame-dataset-aerial-imagery-pile-burn-detection-using-drones-uavs}    \\[2mm]

\myrowcolour
16    &FF-Det      &\href{https://data.mendeley.com/datasets/gjmr63rz2r/1}{https://data.mendeley.com/datasets/gjmr63rz2r/1}    \\[2mm]

17    &FireNet      &\href{https://github.com/OlafenwaMoses/FireNet}{https://github.com/OlafenwaMoses/FireNet}    \\[2mm]

\myrowcolour
18    &AIDER      &\href{https://github.com/ckyrkou/AIDER}{https://github.com/ckyrkou/AIDER}    \\[2mm]

19    &ForestryImage      &\href{https://www.forestryimages.org/browse/subthumb.cfm?sub=740}{https://www.forestryimages.org/browse/subthumb.cfm?sub=740}    \\[2mm]

\myrowcolour
20    &FIRE      &\href{https://www.kaggle.com/datasets/phylake1337/fire-dataset}{https://www.kaggle.com/datasets/phylake1337/fire-dataset}    \\[2mm]

21    &CAIR      &\href{https://github.com/cair/Fire-Detection-Image-Dataset}{https://github.com/cair/Fire-Detection-Image-Dataset}    \\[2mm]

\myrowcolour
22    &FESB MLID      &\href{http://wildfire.fesb.hr/}{http://wildfire.fesb.hr/}    \\[2mm]

23    &FiSmo      &\href{https://github.com/mtcazzolato/dsw2017}{https://github.com/mtcazzolato/dsw2017}    \\[2mm]

\myrowcolour
24    &Corsican      &\href{https://cfdb.univ-corse.fr/index.php}{https://cfdb.univ-corse.fr/index.php}    \\[2mm]

25    &FireSense      &\href{https://zenodo.org/records/836749}{https://zenodo.org/records/836749}    \\[2mm]

\myrowcolour
26    &BoWFire      &\href{https://bitbucket.org/gbdi/bowfire-dataset/downloads/}{https://bitbucket.org/gbdi/bowfire-dataset/downloads/}    \\[2mm]

27    &MIVIA      &\href{https://mivia.unisa.it/datasets/video-analysis-datasets/fire-detection-dataset/}{https://mivia.unisa.it/datasets/video-analysis-datasets/fire-detection-dataset/}    \\[2mm]

\myrowcolour
28    &VisiFire      &\href{http://signal.ee.bilkent.edu.tr/VisiFire/}{http://signal.ee.bilkent.edu.tr/VisiFire/}    \\[2mm]

29    &FireClips      &\href{http://signal.ee.bilkent.edu.tr/VisiFire/Demo/FireClips/}{http://signal.ee.bilkent.edu.tr/VisiFire/Demo/FireClips/}    \\[2mm]
\bottomrule
\end{tabular}}
\end{center}
\footnotesize{\vspace{0.5mm}
}
\end{table*}

\begin{figure}[H]
    \centering
    \centerline{\includegraphics[width=1\textwidth]
    {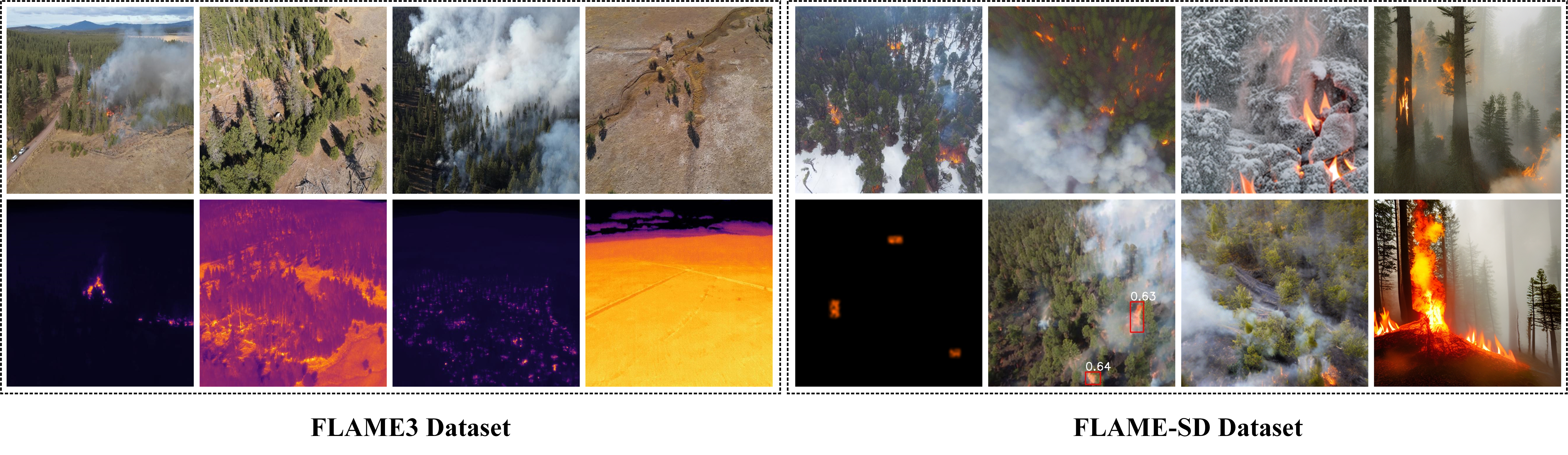}}
    \caption{Representative samples from forest fire datasets: FLAME3 dataset and FLAME-SD dataset.}
    \label{fig: zero}
\end{figure}

The \textbf{FLAME3} dataset is the latest addition to the Flame collection of wildfire images (see FLAME1 and FLAME2 for more information on those specific datasets) \cite{Flame3_dataset}. The images in this dataset were collected between 2022 and 2023 with a DJI M30T UAV from various locations, including TNC Sycan Marsh Preserve, Oregon, USA; San Carlos Apache Tribe, Arizona, USA; Tonto National Forest, Arizona, USA; Coconino National Forest, Arizona, USA, and Tall Timbers Research Station, Florida, USA. As shown in Figure \ref{fig: zero}, the FLAME3 dataset includes synchronized image quartet blocks: raw thermal JPEG, radiometric thermal TIFF, corrected field-of-view RGB, and raw RGB images. This dataset represents a crucial advancement for AI-driven aerial wildfire management by integrating visual spectrum and radiometric thermal imagery for comprehensive analysis. The resolution of the FLAME3 images is delivered in $4000p\times3000p$ for the raw RGB images, while the thermal images are presented in a lower resolution of $640p\times512p$. The dataset repository is classified into two categories: images containing burning fire (labeled as “fire”) and images depicting forest environments without fire (labeled as “no-fire”). Images labeled as "fire" consist of 622 sets of raw RGB, raw thermal, corrected RGB, and radiometric thermal TIFF images. Similarly, images labeled as “no-fire” comprise 116 sets of raw RGB, raw thermal, corrected RGB, and radiometric thermal TIFF images, resulting in a total of 2,952 images. The images in this dataset were annotated via radiometric thermal TIFF to reduce the time and error typically involved in labeling these datasets by hand. Thermal TIFF works by assigning a temperature threshold to each pixel in an image, allowing for the binary classification of fire vs. no fire to be determined by the temperature assigned to the pixel, with high temperatures being associated with fire. The limitation of the FLAME2 dataset was the misalignment between RGB and thermal image pairs. To address this issue, this work focused on ensuring near pixel-perfect alignment between the image pairs, with efforts taken to fully resolve the alignment challenges. The overall aim of this dataset was to continue where FLAME1 and FLAME2 left off, introducing new data-driven by the addition of the thermal TIFFs that help to provide researchers with a novel and accurate way to assess fire in a forest setting \cite{ONeill2024PixelsTP}. This dataset is licensed under the MIT license.

The \textbf{FLAME-SD} dataset is a collection of synthetic wildfire images created by the AI-SENDS lab at Clemson University, USA. The data is generated through an artificial process using diffusion models that combine real wildfire images with synthesized fire masks, as illustrated in Figure \ref{fig: zero}. The dataset is mainly built upon the FLAME1, FLAME2, and D-Fire datasets. The generating process started by extracting RGB values from actual flames in the D-Fire dataset and fusing them with randomly generated binary masks. These masks are refined with Perlin noise, producing smoother and more naturalistic patterns. Then, the generated masks fused with the original wildfire images from the FLAME1 and FLAME2 datasets to form synthesized data. Afterward, the images are processed through a pre-trained Variational Autoencoder (VAE) to encode them into latent representations. A diffusion model (SDv1.5) denoises these representations, guided by a text prompt that specifies the desired output. Finally, the denoised latent tensors are decoded back into images by the VAE to produce realistic wildfire images as well as their corresponding ground-truth masks. The dataset contains a total of 10,000 images with 4K resolution, including a subset that consists of frame-mask pairs. This dataset is generated to train and test fire classification aimed at classifying images as either containing fire or not. Additionally, it is suitable for various testing and training options for object detection models to identify fire.

\begin{figure}[H]
    \centering
    \centerline{\includegraphics[width=1\textwidth]
    {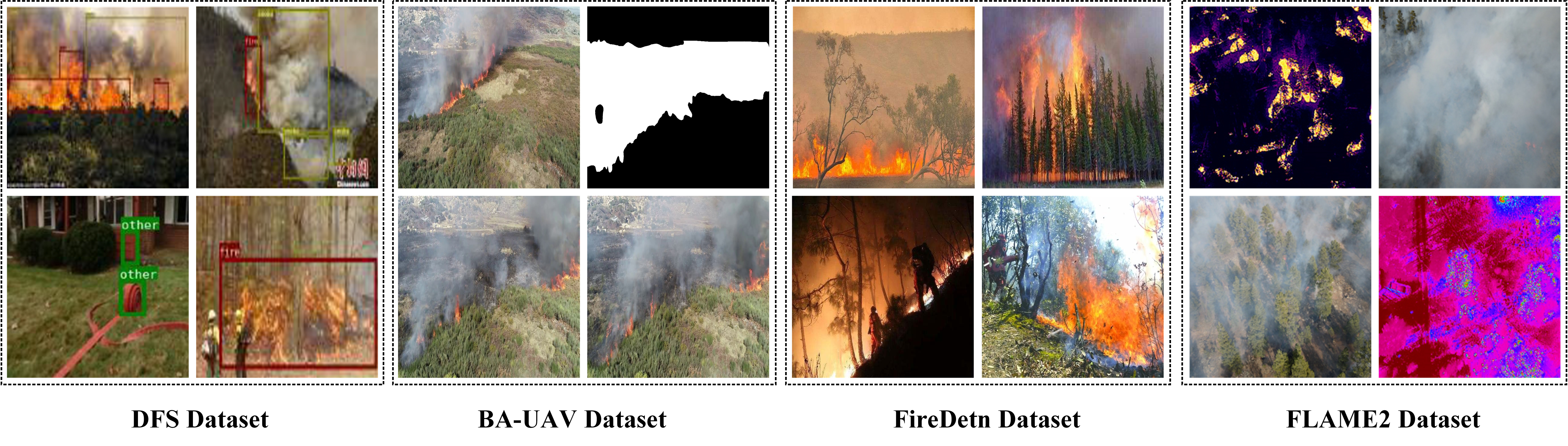}}
    \caption{Representative samples from forest fire datasets: DFS dataset, BA-UAV dataset, FireDetn dataset, and FLAME2 dataset.}
    \label{fig: first}
\end{figure}

The \textbf{DFS} dataset compiled by Siyuan Wu, Xinrong Zhang, Ruqi Liu, and Binhai Li, consists of high-quality images and annotations of “fire” and “smoke”, with objects that could be mistaken for fire in color and brightness being classified as “other”. Figure \ref{fig: first} demonstrates some examples of the DFS dataset, showcasing its diversity and the variety of scenarios it encompasses. This classification prevents models from being confused and training on objects they incorrectly identify as fire. The dataset contains 9,462 total images, allowing for various testing and training options for object detection models to identify fire. A unique aspect of this dataset is its heavy focus on “fire,” “smoke,” and “other” annotations. The “other” category helps object detection models distinguish true fire from similar objects, such as streetlights. Since “fire” and “smoke” often appear together, emphasizing smoke detection helps the model understand their high correlation. The dataset is split into three categories based on the size of the annotated fire: large, medium, and small. This allows models to be trained to detect various kinds of fire, from raging forest fires to the flame produced by a lighter. The DFS dataset is a compilation of images from Baidu, Google Images, Sharma, Dunnings, and Leilei, ensuring a diverse range of fire images. It is further subdivided into indoor and outdoor fire categories. Examples of outdoor fires include factory fires, buildings on fire, and burning cars, while examples of indoor fires include electrical fires and cooking flames. The images were annotated using LabelMe and converted into PASCAL VOC, COCO, and YOLO file formats. Performance metrics such as precision and recall are used to evaluate detection performance. IoU (Intersection over Union) measures how much the predicted bounding box overlaps with the desired part of the image. Finally, the mAP (mean Average Precision) measures the precision for each class, averages the values for all classes, and returns the final mean score. A limitation of this dataset is the data imbalance caused by the ‘other’ category. Since fire is the dataset’s primary focus, there are far more ‘fire’ instances than ‘other’ instances. This imbalance is reflected in the mAP50 scores: 62.29\% for fire (based on 2,059 fire images) versus 34.41\% for “other” (based on 1,034 images). However, when only the fire detector is trained, the mAP50 falls to 45.73\%. This demonstrates that the “other” category can sometimes be misclassified as fire, highlighting a bias that must be addressed when training fire detection models on datasets containing the “other” class.

The \textbf{BA-UAV} dataset is a comprehensive collection of wildfire frames and segmentation masks compiled by the CIIT C\&T Research Centre at the Polytechnic University of Leiria. The video that yielded the frames was recorded via a UAV at a resolution of $720p\times1280p$, surveying an area burned by a forest fire in northern Portugal. This area predominantly consists of granite-rich soil and is primarily covered by undergrowth shrubs, with some maritime pine and oak trees scattered throughout. The dataset contains frames annotated for wildfire segmentation, as provided in Figure \ref{fig: first}, serving as a basis for model training and performance evaluation. It consists of 249 frames and segmentation masks, with the training subset comprising 226 frame-mask pairs and the testing subset containing 23 pairs. The training and validation frames were created by sampling every fourth second, resulting in a selection rate of 1 out of every 100 frames, with sampling stopping at 22,500 frames. The testing dataset’s sample rate, also 1 out of 100 frames, is offset by two seconds (or 50 frames) to prevent overlap with the training and validation frames. This ensures that the testing frames represent unseen data, enabling fair and unbiased evaluations. The frames were annotated using the Labelme image annotation tool, which allowed the team to draw segmentation polygons using a computer mouse. Frames were divided into two classes: ``burned-area’’ representing the landscape affected by the fire, and ``unburned-area’’ representing regions untouched by the fire. The segmentation models used include U-NET Base, U-Net Red architectures, and U-Net 3D architecture. A noted limitation of this dataset is its weak generalization, as further enrichment and diversity are needed to improve predictions for fire in new videos. The dataset aims to provide researchers with a high-quality, segmented collection of wildfire-affected frames to address the lack of available datasets for this purpose.

The \textbf{FireDetn} dataset was created by the team of Xiaotian Wang, Zhongjie Pan, Hang Gao, Ningxin He, and Tiegang Gao for the purpose of providing researchers with wildfire-specific images. The dataset offers a streamlined dataset for training, validating, and testing a wildfire detection model, as shown in Figure \ref{fig: first}. The number of images in this dataset is as follows: the training set contains 2,806 images, the validation set contains 934 images, and the testing set contains 934 images. The resolution of the images used in the FireDetn dataset is $640p\times640p$.  The FireDetn dataset is annotated via bounding boxes around the fire areas of the images to make them suitable for wildfire detection models. The images are broken down into different flame sizes, with four detection heads, ranging from small flames occupying a shallow feature map to large flames that hold a lot of information, which require a deeper feature map. The team behind this dataset felt strongly about having the dataset be composed entirely of forest fires to ensure that models could train accurately on forest fire scenarios. The idea behind this is that researchers can train models to precisely and accurately detect forest fires in real-time situations. The metrics used to evaluate the dataset are AP50, AP75, AP, APs(small), APm (medium), APl (large), AR1 (1 detection per image), AR10 (10 detections per image), and FPS. AP stands for average precision and yields the area under the precision-recall curve, while AR is average recall, which shows the intersection over union, indicating the accuracy of the object detection. FPS is the mean number of frames the model can detect per second, which indicates the model’s speed. A limitation of this dataset is the lack of diversity in the set, containing images of forest fires in ideal weather conditions, which need to be expanded upon in future iterations.

The \textbf{FLAME2} dataset is a comprehensive collection of fire images and videos developed by IS-WiN Lab at Clemson University, USA \cite{FLAME2Dataset}. This dataset has been specifically designed for fire detection and segmentation tasks, as depicted in Figure \ref{fig: first}. The FLAME2 dataset includes 53,451 RGB/IR frame pairs parsed from videos from 7 RGB and IR video sequences captured in an open canopy in the Kaibab National Forest in Arizona in November 2021, capturing pinyon-juniper woodlands and ponderosa pine forests. The data has resolutions up to $3840p\times2160p$ for the RGB video and $640p\times512p$ for the IR video, with down-scaling to $254p\times254p$, which is the desired resolution for the CNN model that is being used to detect if fire or smoke is present in the provided frame. Annotations providing bounding boxes and segmentation masks for fire regions, alongside metadata on fire size and severity, to classify whether part of the frame is ``fire’’ or ``no-fire’’ and if the frame is ``smoke’’ or ``no-smoke’’ if smoke fills more than 50\% of the frame. Each individual frame pair has been reviewed by two experts by hand. A unique feature of the FLAME2 dataset is the combination of infrared and RGB images, which allows for a greater perspective on fire detection. The dataset has been used in object detection and segmentation benchmarks, with meticulously calculated metrics like recall, precision, and the F1-Score used to evaluate the model’s performance on the dataset. The FLAME2 dataset is publicly accessible under the Creative Commons Attribution (CC BY) license, though limitations, such as the specific environment of the dataset, could pose a problem should a dramatically different environment, such as an urban building, be shown instead. Researchers have leveraged this dataset for early fire detection systems, demonstrating its pivotal role in advancing wildfire management technologies \cite{10633236,10620877,10226139,Meleti}. FLAME2’s supplemental dataset provides details and additional variables to enhance the main dataset. The supplemental dataset allows for a superior ability to detect fire and train models based on this dataset. This dataset includes various weather conditions, burn plans, RGB pre-burn point clouds, raw pre-burn videos, and other information that can be used to fine-tune a model’s accuracy. Using the supplemental dataset after training and testing a model on the main dataset can increase and enhance a model’s accuracy and performance, such as training on a pre-burn frame of the area, allowing the model to have a better understanding of fire and when it occurs based on the “before and after” comparison it is now making. Weather data can be used not only to understand what conditions lead to fire occurring but also to allow for a better understanding of details surrounding a fire, which can enhance its predictive capabilities. Limitations of this supplemental dataset are similar to those shown in the main dataset, with the niche area that the video and frames are derived from, Northern Arizona, which has specific flora, weather patterns, and climatology that may not translate well to other regions where a fire may occur. Researchers can use the additional information provided by the supplemental dataset to enhance their model’s performance based on the information provided by the main dataset.

\begin{figure}[H]
    \centering
    \centerline{\includegraphics[width=1\textwidth]
    {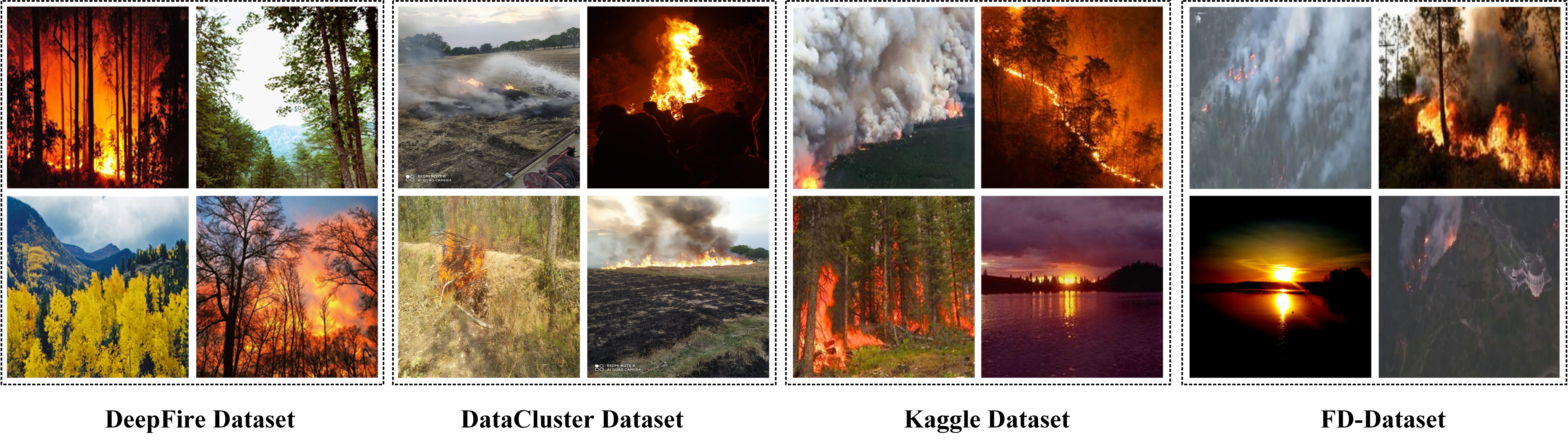}}
    \caption{Representative samples from forest fire datasets: DeepFire dataset, DataCluster dataset, Kaggle dataset, and FD-dataset.}
    \label{fig: second}
\end{figure}

The \textbf{DeepFire} dataset is a binary fire detection image dataset comprising forest landscapes. It was created by Ali Khan, Bilal Hassan, Somaiya Khan, Ramsha Ahmed, and Adnan Abusassba and consisted of 1,900 images evenly divided into two classes: “fire” and “no-fire.” This dataset is designed to train and test fire detection models for binary classification, as shown in Figure \ref{fig: second}. The images are three-channeled and have a resolution of $250p\times250p$. The dataset was compiled from a variety of online sources using keywords such as “mountain fire,” “forest fire,” “forest,” and “mountain,” ensuring adherence to the fire/no-fire dichotomy intended by the researchers. Additionally, various performance metrics were used to validate the dataset's quality, including Accuracy, Precision, Recall, Error Rate (ER), and F1-Score. These metrics are based on True Positive (TP), True Negative (TN), False Positive (FP), and False Negative (FN). TP refers to fire images that are correctly classified as fire, while TN represents non-fire images correctly classified as non-fire. Conversely, FP occurs when non-fire images are misclassified as fire, and FN happens when fire images are misclassified as non-fire. Despite the dataset's strengths, it has some limitations, such as false alarms and relatively low spatial resolution. Addressing these issues could enhance its utility for drone-based surveillance systems, enabling early forest fire detection. The primary aim of this dataset is to provide researchers with a high-quality resource for training fire detection models capable of accurately distinguishing fire from non-fire images. This would greatly support the development of automated surveillance systems that can detect forest fires at an early stage, alerting the proper authorities so that the fires can be contained and eliminated before they pose a significant ecological and humanitarian threat.

The \textbf{DataCluster} dataset compiled by DataCluster labs is a collection of fire and smoke images from a wide range of sources, both captured and collected from rural and urban areas. This feature enable the dataset to provides a diverse set of images to train and test various machine learning algorithms. Figure \ref{fig: second} represents some instances from the DataCluster dataset. This dataset contains almost 7,000 images collected from around 400 cities in India using mobile phones from the time-frame of 2020-2021. All the images are presented in HD format, with a minimum resolution of $1920p\times1080p$. These images were annotated in a variety of formats, including COCO, YOLO, PASCAL-VOC, and TF-Record. This dataset is licensed under the copyright of the original owners, DataCluster labs, for lease for research and commercial purposes. These images contain a wide range of conditions, such as day, night, close and far distances, and vantage points, allowing models to experience a broad spectrum of information, putting their accuracy to the test. The limitation of this dataset is its broad scope, which may lead a model trained on it to to become versatile in detecting fires but being less proficient at distinguishing between specific scenarios, such as forest fires and urban fires. The purpose of this dataset is to give researchers the ability to train a model on a wide range of fire and smoke images with a broad spectrum of settings.

The \textbf{Kaggle} dataset, created by Baris Dincer, is a collection of wildfire images, some of which contain fire, while others do not, as illustrated in Figure \ref{fig: second}. This dataset is publicly available on Kaggle for the purpose of giving researchers a testing and training/validation set to improve their model’s performance. There are a total of 1,900 images in the Kaggle dataset, provided in .JPG format. This dataset is a binary classification system, with a diverse set of images that either do or do not contain fire with “Fire” and “No-Fire” annotations. The main objective of this dataset is to provide a wide range of forest fire images for training and testing various detection models, enabling them to become more efficient and accurate at identifying forest fires in real-time. The main drawback of this dataset is its limited scope, as models trained on forest fire images may experience a substantial decrease in detection accuracy when encountering non-forest fire images. This dataset is licensed by Open Data Commons under the Database Contents License (DbCL).

The \textbf{FD-dataset} combines two datasets: BowFire and Foggia \cite{foggia2015real} datasets, encompassing 31 videos, where 14 videos depict fire scenarios and 17 videos show non-fire situations. Additionally, the dataset contains a collection of fire and non-fire images sourced from the internet. This comprehensive dataset includes 50,000 images of varying resolutions, equally divided between fire-related and non-fire-related scenarios. Some selected images are represented in Figure \ref{fig: second}. The fire-related images capture diverse scenarios such as flame, smoke, and burn effects, while the non-fire images feature fire-like objects, including sunsets and sunrises, offering challenging cases for classification. The dataset spans both forested and non-forested environments.

\begin{figure}[H]
    \centering
    \centerline{\includegraphics[width=1\textwidth]
    {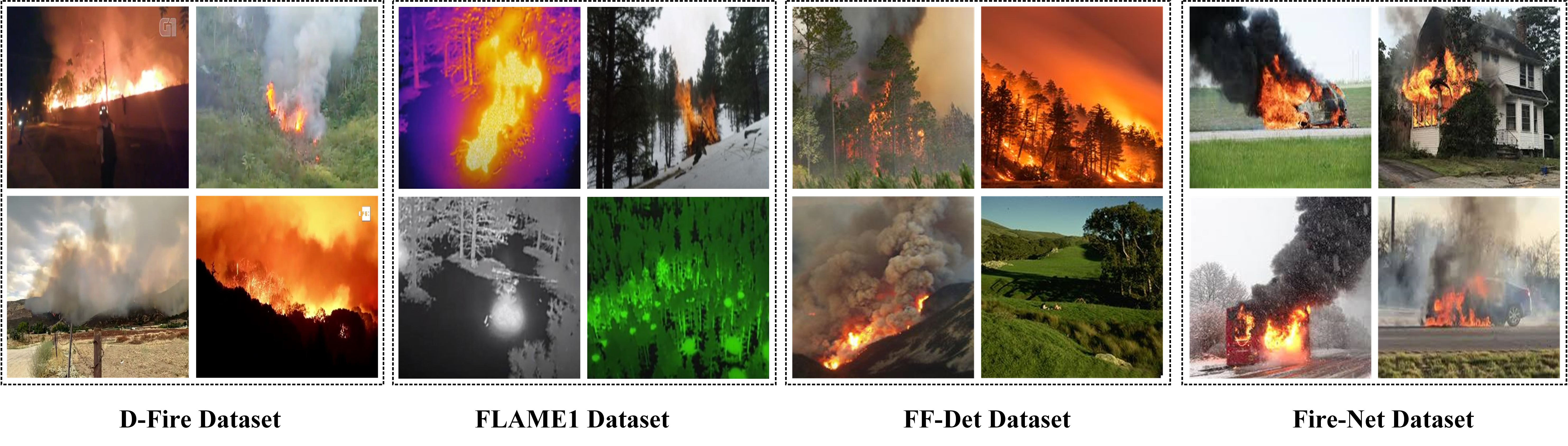}}
    \caption{Representative samples from forest fire datasets: D-Fire dataset, FLAME1 dataset, FF-Det dataset, and Fire-Net dataset.}
    \label{fig: third}
\end{figure}

The \textbf{D-Fire} dataset is a collection of fire and smoke images, as well as surveillance video footage, gathered by researchers from GAIA. This dataset is principally designed for machine learning object detection algorithms, as shown in Figure \ref{fig: third}. This dataset contains 21,527 images, summarized as follows: the fire category contains 1,164 images, the smoke category contains 5,867 images, the fire and smoke category contains 4,658 images, and the none category (images that do not contain fire or smoke, but have objects or environments that could be mistaken for smoke or fire) contains 9,838 images. The RGB images used in this dataset have a resolution of $416p\times416p$. These images were annotated with bounding boxes to indicate whether they contain fire or smoke, clearly showing the regions of interest. In total, there are 26,557 bounding boxes, with 11,865 being labeled as “smoke” and 14,692 labeled as “fire”. The images used in this dataset are collected from various sources to give it a broad base to train machine learning models. The sources for the images are from legal fire simulations in the Technological Park of Belo Horizonte, Brazil, surveillance camera footage from landscapes at Universidade Federal de Minas Gerais (UFMG) and at Serra Verde State Park in Belo Horizonte, as well as images collected from the Internet. The images in this dataset contain a range of conditions to allow for better training and testing packages for a model using this dataset, with a multitude of angles, bright light and darkness, and a host of different environments. For the surveillance footage, with insects landing on the camera lens, as well as filming adverse weather conditions like rain, which allow models to train on an extremely dynamic dataset. The performance metrics used for this dataset are mAP50, AP50 smoke, AP50 fire, the F1-Score, and the average IoU. The images in this dataset belong to the public domain, while the image collection resulting from the team’s work is made available through the Creative Commons Zero v1.0 Universal License.

The \textbf{FLAME1} dataset (an acronym standing for Fire Luminosity Airborne-based Machine Learning Evaluation) is a collection of videos and images gathered by aerial drones (UAV) in the ponderosa pine forests in Observatory Mesa in northern Arizona \cite{qad6-r683-20}. Figure \ref{fig: third} illustrates some selected images from the FLAME1 dataset. This dataset is annotated with binary labels of “fire” and “no-fire” for a model to be trained on object classification. It also contains pixel-wise masks for segmentation tasks, allowing for fire expansion modeling. There are 39,375 frames in the training dataset that are labeled for fire classification, with 25,018 frames for fire and 14,357 frames for non-fire. In the testing dataset, there are 8,617 frames, with 5,137 labeled for fire and 3,480 labeled for non-fire. In addition, there are 2,003 segmentation masks that were manually labeled for fire. The raw data was captured in .MP4 and .MOV formats, and the analyzed format was presented in .JPEG and .PNG. The videos were captured in four formats, normal light spectrum, fusion, white-hot, and green-hot palettes. The resolution for the classification images is $254p\times254p$, while the resolution for the segmentation frames is $3480p\times2160p$. The performance metrics for this dataset are a confusion matrix and accuracy for classification purposes, while precision, recall, AuC (area under the curve), IoU (intersection over union), and F1-score are used for segmentation evaluation. Limitations in this dataset are a lack of environmental diversity in the dataset, as fire can spread rapidly, so having a model be able to detect fire in a multitude of settings would be extremely beneficial for future applications. This dataset is licensed under the GNU General Public License.

The \textbf{FF-Det} dataset is created by Ali Khan and Bilal Hassan to provide a way to tackle the important problem of detecting forest fires. Using a multitude of search engines, the team behind this dataset was able to retrieve the images used for this dataset. All the collected images has a resolution of $250p\times250p$ and are three-channeled, as demonstrated in Figure \ref{fig: third}. Each image was meticulously reviewed and edited by a team member to remove unnecessary elements, such as people, fire suppression equipment, and cropped to ensure the main focus was on the fire. This dataset contains 1,900 total images, with 950 belonging to the “fire” class and 950 belonging to the “no-fire” class. The dataset was divided into an 80:20 split, meaning that 380 images are for testing and 1,520 are for training. This dataset is designed to have a binary classification for fire and no-fire in an image so that researchers can train a model to detect fire using these images. This dataset is licensed under a Creative Commons Attribution 4.0 International License.

The \textbf{FireNET} dataset is designed to train a model to detect fire in real time to improve the speed and accuracy of fire detection. This dataset was created by Moses Olafenwa to provide a diverse set of fire images for models to train and test on a wide range of situations. Figure \ref{fig: third} presents the selected samples from FireNet dataset. The images are annotated to identify parts of the image that contain fire. The dataset comprises 502 images, with 90 for validation and 412 for training. The images in this dataset have a resolution of $480p\times480p$. This dataset is licensed under the MIT License.

\begin{figure}[H]
    \centering
    \centerline{\includegraphics[width=1\textwidth]
    {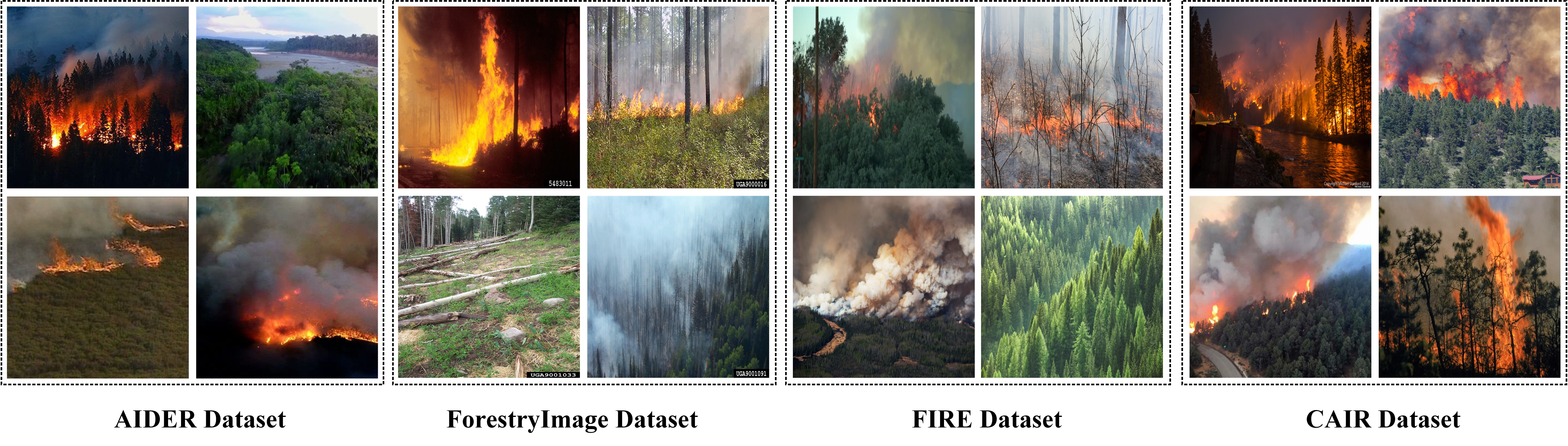}}
    \caption{Representative samples from forest fire datasets: AIDER dataset, ForestryImage dataset, FIRE dataset, and CAIR dataset.}
    \label{fig: fourth}
\end{figure}

The \textbf{AIDER} (Aerial Image Dataset for Emergency Response) is a database comprised of subsets: collapsed building/rubble, fire/smoke, flood, traffic accidents, and normal, with the fire/smoke dataset being the focus of this research. This dataset has 8,540 images overall, but has 740 total fire/smoke images, with 420 being used for training, 110 being used for validation, and 210 being used for testing. Figure \ref{fig: fourth} demonstrates some examples of the AIDER dataset, showcasing its diversity and the variety of scenarios it encompasses. These images were collected from online sources, like Google Images, Bing, and other sites, for example, using keywords, UAV, drone, and aerial view. The images for fire and non-fire are composed of a $128p\times128p$ resolution. Annotations for these images are done by applying labels to the appropriate hazard, in this case, labeling “fire” and “no-fire”. These images contain a wide range of settings and environments to allow the model to recognize fire in a variety of environments. The performance metrics used to evaluate this dataset were the Frames-Per-Second (FPS) and mean F1-Score. This dataset is licensed under the GNU General Public License version 3.

The \textbf{ForestryImage} dataset is a joint collaboration between the University of Georgia’s Center for Invasive Species and Ecosystem Health, the US Forest Service, the International Society of Arboriculture, and the USDA Identification Technology Program seeking to enhance education and research concerning forests. The images capture a broad range of flora and fauna found in forest environments, seeking to promote a better understanding and appreciation of forest ecosystems. Each image comes with information and a name, such as fire for images that contain fire in them. There are 730 fire-labeled images in this dataset, as shown in Figure \ref{fig: fourth}. The resolution of these images varies based on composition, but they are roughly $768p\times511p$. These images are licensed under many different licenses based on the photographer who captured the image.

The \textbf{FIRE} dataset was created by the NASA Space Apps Challenge in 2018. The primary goal of this dataset is to provide a robust foundation for developing machine learning models capable of accurately recognizing images containing fire. Figure \ref{fig: fourth} visually displays some selected samples from the FIRE dataset. It is specifically designed for binary classification tasks aimed at classifying images as either containing fire or not. The dataset is generally organized into two categories: the “fire” subset contains 755 samples of images with fire and/or heavy smoke, while the “non-fire” subset contains 244 images of diverse nature scenes containing mostly trees, forest, lake, waterfall, and foggy forest. Also, the images are captured under different environmental conditions, such as varying levels of brightness, smoke density, and flame intensity. All samples are in PNG format containing RGB images of varying resolutions. The data distribution is imbalanced, necessitating careful processing and partitioning to ensure reliable results. It should be noted that the dataset contains no annotation details other than its fire and non-fire folders. Although the FIRE dataset is a valuable dataset for fire classification tasks, it has several limitations that affect its applicability. The first challenge with this dataset is its insufficient size and diversity, which may lead to overfitting issues. Another challenge is the possible class imbalance, where an unequal number of fire and non-fire images could result in biased model predictions. Furthermore, the dataset may contain challenges in fire-like scenarios, such as sunsets or red objects, which increase the likelihood of false positives if these cases are not well-represented.

The Fire-Detection-Image-Dataset uploaded by Github user \textbf{CAIR} is designed to show images of fire and normal surroundings of a multitude of areas, with an emphasis on various fire situations, like light, size, brightness, and the place where the fire is happening. The images are annotated as containing fire or not containing fire. These images are in .rar format, with one batch containing the fire images and five batches containing normal images. Some examples of the CAIR dataset are shown in Figure \ref{fig: fourth}.

\begin{figure}[H]
    \centering
    \centerline{\includegraphics[width=1\textwidth]
    {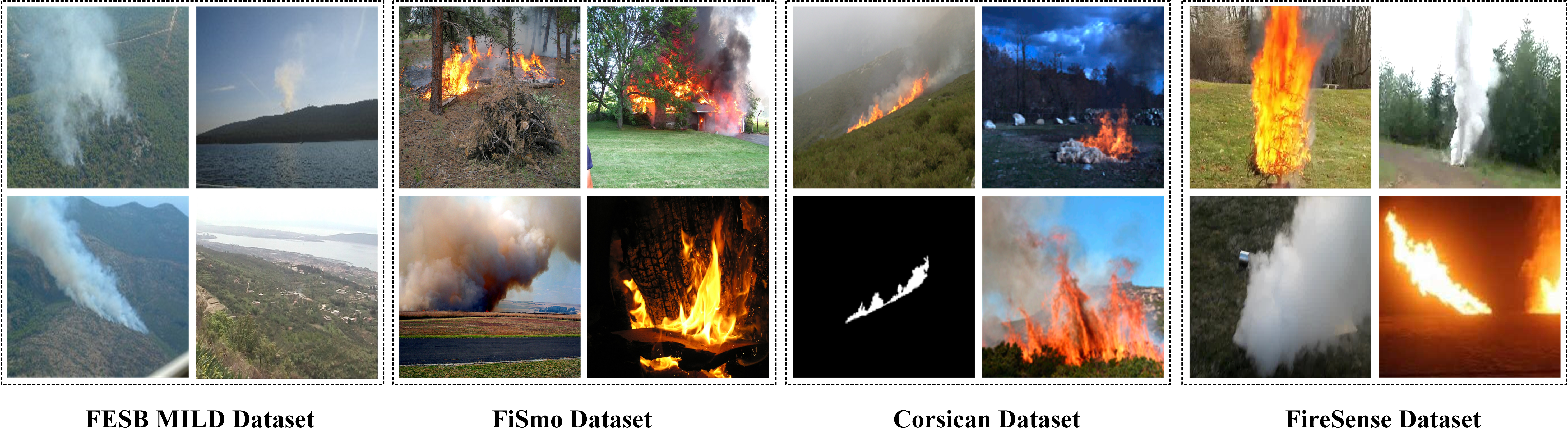}}
    \caption{Representative samples from forest fire datasets: FESB MILD dataset, FiSmo dataset, Corsican dataset, and FireSense dataset.}
    \label{fig: fifth}
\end{figure}

The \textbf{FESB MLID} dataset was developed by the Faculty of Electrical Engineering and Naval Architecture (FESB) under the Wildfire Research Center at the University of Split, Croatia. This dataset comprises 400 Mediterranean landscape images paired with their corresponding semantic segmentation maps that are meticulously hand-labeled to use as ground truth images, as presented in Figure \ref{fig: fifth}. FESB MLID dataset contains 11 segmentation classes, such as smoke, clouds, rocks, fog, sunlight, sky, water surface, vegetation, landscape, and others. Each sample consists of a $768p\times576p$ resolution RGB image and has an equally sized bitmap file for its segmentation. Notably, the dataset includes many examples of distant smoke features, enhancing its utility for smoke detection tasks. Although the dataset includes many examples of clear skies, a good portion of the dataset is smokey, foggy, or contains many clouds, making it valuable for developing and evaluating algorithms for the segmentation and classification tasks, particularly in the context of automated wildfire monitoring systems. By providing a diverse set of Mediterranean landscape images with detailed segmentation information, the FESB MLID dataset will provide an excellent contribution to the development and testing of algorithms in order to improve wildfire detection and environmental analysis. Scientists can access the FESB MLID dataset through the Wildfire Observers and Smoke Recognition portal, which provides them with a collaborative platform for advancements in automated smoke detection processes.

The \textbf{FiSmo} dataset is a compilation of datasets from emergency situations, containing both images and videos of various fire and smoke events. It was developed as part of the RESCUER Project by the Databases and Images Group (GBdI) at the University of São Paulo in São Carlos, Brazil. FiSmo images are presented in Figure \ref{fig: fifth}, which are a combination of the Flickr-FireSmoke, Flickr-Fire, BoWFire, and SmokeBlock datasets. The Flickr-FireSmoke dataset consists of 5,556 images with the purpose of fire and smoke detection. After annotation, the dataset was categorized into four classes: “fire and smoke,” “only fire,” “only smoke,” and “none.” There were 527 fire and smoke images, 1,077 only fire images, 369 only smoke images, and 3,583 images with none. The Flickr-Fire dataset contains 2,000 images for global fire detection and content-based image retrieval. Of these, 1,000 images are of fires, and 1,000 are not fires. The dataset also has six files with low-level features extracted from the images. These features were extracted based on color layout, scalable color, color structure, color temperature, edge histogram, and texture browsing. The BoWFire dataset contains 226 images for fire detection and segmentation. Of these, 119 images are of fire, and the other 107 are not of fire. The dataset also contains 226 masks of the fire regions and 240 Regions of Interest (ROIs), which are $50p\times50p$; 80 of these are labeled as fire and 160 as not fire. Lastly, the SmokeBlock dataset contains 1,666 images for smoke detection and segmentation. Of these, 832 images are labeled as smoke, and the other 834 are labeled as not smoke. It also contains 10 files with low-level features collected from the images, such as color layout, color structure, color temperature, edge histogram, LPB, normalized histogram, scalable color, and texture spectrum. SmokeBlock also contains 103 ROIs, 43 of which are labeled as smoke and 60 as not smoke. FiSmo dataset has 97 videos collected from YouTube, queried by this set of keywords: fire, smoke, explosion, flames, burning, blaze, campfire, bonfire, combustion, ignite, wildfire, and firefighters. These videos are annotated frame by frame with the labels fire, not fire, and ignore. FiSmo is publicly accessible under the Creative Commons license. It plays a pivotal role in emergency image and video analysis tasks, specifically in fire and smoke detection, segmentation, and content-based retrieval.

The \textbf{Corsican} fire dataset is a collection of wildlife imagery developed by the Sciences Pour I’Environment laboratory within the University of Corsica. This database was created with the purpose of modeling and experimenting with vegetation fires. It includes 500 visible wildfire images, 100 image pairs combining visible wildfire and near-infrared area, and 5 sequences of image pairs combining visible wildfire and near-infrared area. Figure \ref{fig: fifth} provides a visual representation of the images in the Corsican dataset. The data spans multiple modalities, including both RGB and IR images captured under diverse conditions such as vegetation type, climate, brightness, and distance to fire. Each image is described by 22 features. These features include the spectral range of the image, camera model, sensitivity, exposure time, time of the shot, moment of the day/night, the fire distance from the camera, place, region, GPS position, direction of propagation of the fire, vegetation type, color of smoke, presence of clouds, presence of trucks and men, ground truth, image dimensions, percentage of the image that is fire, dominant color of the flame, texture level of the fire, fire/smoke covering percentage, and environmental brightness. One notable feature is the percentage of fire/smoke covering. This feature measures the ratio of fire pixels with smoke to the total fire pixels. It is then categorized into three ranges: low if the percentage is between 0 and 20, medium if the percentage is between 20 and 45, and high if the percentage is between 45 and 100. Another unique feature is that each image is divided into three categories based on the dominant color of the fire: red, orange, and yellow. Each pixel is classified in one of these colors using the TSL color space, specifically using hue (channel T of the color space) to distinguish the colors. The dataset is publicly accessible upon signing the Corsican Fire Database License Agreement. The license states that the dataset must be used for research purposes only and cannot be redistributed or modified. Researchers have used this database to measure the geometric characteristics of fires, demonstrating its critical role in advancing technologies in wildfire research and firefighter communities.

The \textbf{FireSense} dataset is a collection of 49 videos designed for both flame and smoke detection in RGB format. Figure \ref{fig: fifth} shows sample images from the FireSense dataset. This dataset was developed by the Information Technologies Institute in Greece and Bilkent University in Turkey. There are 11 positive and 16 negative videos for flame detection, while for smoke detection, there are 13 positive and 9 negative videos. This dataset contains data collected from a variety of situations, including what appears to be a forest fire. The other positive videos include scenes such as a fireplace, a fire in a sink, and a few small contained fires outdoors. This dataset has been used in benchmarks for spatio-temporal flame modeling and dynamic texture analysis for video-based fire detection. It is publicly accessible under an Attribution 4.0 International (CC BY 4.0) license. However, its lack of thermal videos and limited diversity in wildfire-specific scenarios may pose challenges for certain applications and should be considered when designing models. Despite these limitations, the FireSense dataset has had a significant role in advancing fire detection technologies.

\begin{figure}[H]
    \centering
    \centerline{\includegraphics[width=1\textwidth]
    {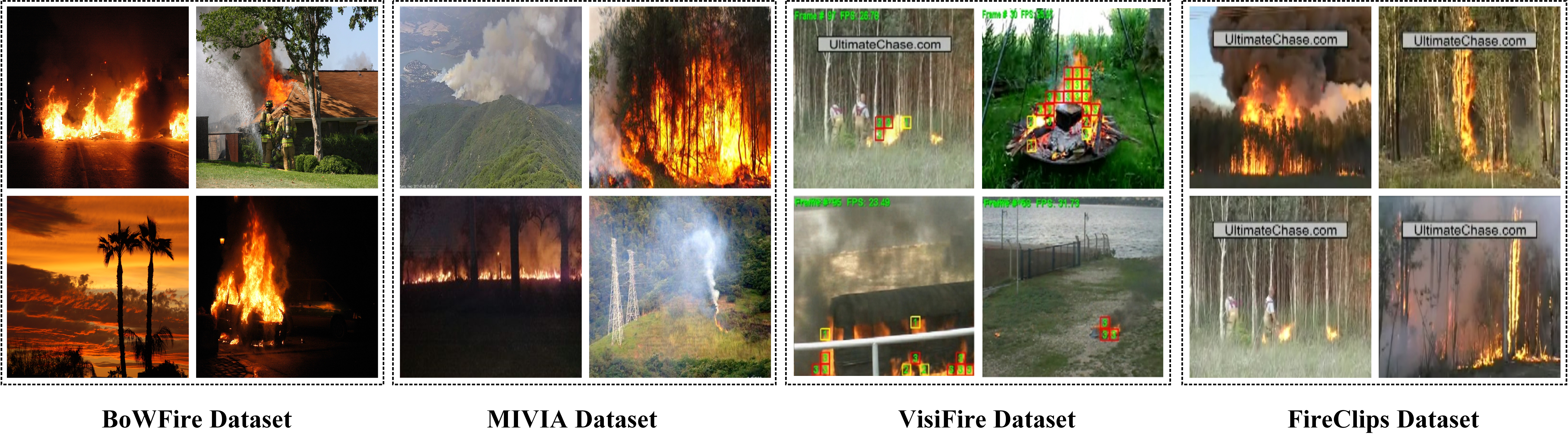}}
    \caption{Representative samples from forest fire datasets: BoWFire dataset, MIVIA dataset, VisiFire dataset, and FireClips dataset.}
    \label{fig: sixth}
\end{figure}

The \textbf{BoWFire} dataset comprises 226 images of different resolutions, mostly in RGB format, along with a training set of 240 images at a resolution of $50p\times50p$. It was created by the Institute of Mathematics and Computer Science at the University of São Paulo. The dataset features a variety of emergency situations, including building fires, industrial fires, car accidents, and riots, as presented in Figure \ref{fig: sixth}. It also includes non-fire images, such as emergency scenes without visible flames, as well as images with fire-like elements, such as sunsets or red and yellow objects. It has been used for classification based on a combination of color and texture information to reduce the number of false positives generated by algorithms. By including diverse emergency scenarios, it captures real-world fire dynamics. It has been used in object detection and segmentation benchmarks with established metrics such as pixel-level classification. It is accessible under an MIT license; however, its small size and limited diversity of wildfire-specific scenarios pose a challenge for certain applications. Despite these limitations, the BoWFire dataset has played a significant role in advancing fire detection technologies.

The \textbf{MIVIA} wildfire dataset is a video collection designed for diverse applications. It includes 31 videos, consisting of 17 fire-related and 14 non-fire-related recordings, gathered from real-world scenarios and also the FireSense dataset. This dataset offers a valuable resource for developing and evaluating fire classification algorithms. Figure \ref{fig: sixth} showcases sample frames from the MIVIA fire detection dataset.

The \textbf{VisiFire} dataset was created by Professor A. Enis Cetin for the purpose of analyzing videos captured via surveillance camera to detect and report fire accurately in real-time. Areas included in the video surveillance footage are outdoor fires, forest fires, and indoor fires, as shown in Figure \ref{fig: sixth}. The dataset contains 12 videos available in resolutions of $400p\times256p$ and $320p\times240p$. The files are accessible in .avi format for researchers to use in their research. This data is licensed under the Fire\&Smoke Detector License. The goal is to detect fire in large and open spaces, which are readily available in video format. These videos exist and are usable to the public under a public domain license.

The \textbf{FireClips} dataset, introduced in 2006, comprises terrestrial video data available in .AVI format. It contains a total of 13 video clips that are categorized into fire and non-fire scenarios, as illustrated in Figure \ref{fig: sixth}. The dataset is captured using camera sensors and primarily focuses on rural and wildfire environments. This collection is valuable for research in wildfire detection and monitoring systems.

Ultimately, Tables \ref{Table: proscons} and \ref{Table: proscons2} highlight the major pros and cons of each dataset, providing a comprehensive comparison to help researchers identify the most suitable dataset for particular applications and objectives.

\begin{sidewaystable*}[htbp]
\centering
\caption{Summary of the strengths and weaknesses of the existing fire and smoke datasets.}
\vspace{1mm}
\label{Table: proscons}
\resizebox{\textwidth}{!}{
\setlength{\tabcolsep}{7pt}
\begin{tabular}{lllllllll}
\toprule
\textbf{No.}    &\textbf{Dataset}   & \textbf{Pros} & \textbf{Cons}  \\
\midrule 

1     
&FLAME3
&\tabitem Includes aligned visual spectrum and thermal images, facilitating fire analysis.
&\tabitem Contains more fire compared to no-fire images, potentially biasing model training.
\\[1mm]
&&\tabitem Data gathered from multiple locations with varied fuel types and burn behaviors.
&\tabitem Thermal images have a lower resolution, which may affect the precise detection.
\\[3mm]

\myrowcolour
2       
&FLAME-SD
&\tabitem Contains 10,000 4K synthesized images, providing rich data for model training.
&\tabitem Being artificially generated, the dataset may lack the complexities of real scenarios.
\\[1mm]
\myrowcolour
&&\tabitem Utilizes noise-based masks for accurate flame placement and synthetic images.
&\tabitem Synthetic backgrounds may lack real-world variety, affecting model robustness.
\\[3mm]

3     
&DFS
&\tabitem Images are categorized by fire size, aiding detection across varying fire intensities.
&\tabitem The fire category has more instances than others, potentially leading to model bias.
\\[1mm]
&&\tabitem Contains images from indoor and outdoor settings, enhancing model robustness.
&\tabitem Lacks video sequences, limiting temporal modeling for fire and smoke detection.
\\[3mm]

\myrowcolour
4       
&BA-UAV
&\tabitem Provides burned and unburned areas masks, enabling precise fire segmentation.
&\tabitem Contains limited annotated frames, possibly insufficient for training DL models.
\\[1mm]
\myrowcolour
&&\tabitem Captures aerial footage from actual fire-affected regions, offering realism insights.
&\tabitem Sourced from single-location data, which limits applicability to diverse scenarios.
\\[3mm]

5     
&FireDetn
&\tabitem Provides precise annotations for fire areas, aiding in accurate object detection.
&\tabitem Uniform resolution of $640p\times640p$ may not capture details in high-resolution data.
\\[1mm]
&&\tabitem Comprises data exclusively of forest fires, ensuring effective model development.
&\tabitem Covers only fire detection and excludes other relevant phenomena such as smoke.
\\[3mm]

\myrowcolour
6       
&FLAME2
&\tabitem Provides paired RGB and Thermal images, enhancing fire analysis capabilities.
&\tabitem Data collected from a single location which may limit model generalizability.
\\[1mm]
\myrowcolour
&&\tabitem Offers high-resolution RGB and IR images and videos, enabling detailed analysis.
&\tabitem Images down-scaled to $254p\times254p$, potentially losing crucial details for training.
\\[3mm]

7     
&DeepFire
&\tabitem Contains an even split between classes, facilitating unbiased binary classification.
&\tabitem Due to the small size of the dataset, it may be insufficient for training DL models.
\\[1mm]
&&\tabitem Compiled from various online sources, ensuring a wide range of fire scenarios.
&\tabitem Absence of metadata such as location and time limits the dataset's applicability.
\\[3mm]

\myrowcolour
8       
&DataCluster
&\tabitem Includes various lighting conditions (day and night), distances, and viewpoints. 
&\tabitem Uneven distribution of fire and smoke could affect model training effectiveness.

\\[1mm]
\myrowcolour
&&\tabitem Offers multiple annotation formats, facilitating flexibility in model development.
&\tabitem Sourced only from urban and rural areas, which limits the dataset's applicability.
\\[3mm]

9    
&Kaggle
&\tabitem Contains an even split between classes, facilitating unbiased binary classification.
&\tabitem Sourcing images from the internet can lead to inconsistencies in image quality. 
\\[1mm]
&&\tabitem Includes a variety range of forest fire images, enhancing model robustness.
&\tabitem The dataset's small size (1,900 images) makes it inadequate for training DL models.
\\[3mm]

\myrowcolour
10       
&FD-Dataset
&\tabitem Covers diverse environments and fire-related phenomena, boosting durability.
&\tabitem Lacks of video sequences, limiting temporal modeling for fire and smoke analysis.
\\[1mm]
\myrowcolour
&&\tabitem Features non-fire images with fire-like characteristics, enhancing model robustness.
&\tabitem Absence of contextual information, such as capture times, locations, and conditions.
\\[3mm]

11   
&D-Fire
&\tabitem Contains fire and smoke bounding boxes, facilitating precise object detection.
&\tabitem High number of images in the none class, needing careful training to avoid bias. 
\\[1mm]
&&\tabitem Rich collection of legal fire simulations, surveillance footage, and internet images.
&\tabitem Uniform resolution of $416p\times416p$ limits fine detection in high-resolution scenarios.
\\[3mm]

\myrowcolour
12       
&FLAME1
&\tabitem Offers binary labels and pixel-wise masks, enabling classification and segmentation.
&\tabitem Data collected exclusively from single forests, limiting the model's generalizability.
\\[1mm]
\myrowcolour
&&\tabitem Includes video captured in multiple formats, enhancing the dataset's applicability.
&\tabitem The fire frames (25,018) outnumber non-fire frames (14,357), risking training bias.
\\[3mm]

13   
&FF-Det
&\tabitem Contains an even split between classes, facilitating unbiased binary classification.
&\tabitem Includes a limited number of images, possibly insufficient for training DL models.
\\[1mm]
&&\tabitem Provides 3-channeled data with a uniform resolution, offering a consistent dataset.
&\tabitem Lack of contextual information, such as capture conditions or geographic locations.
\\[3mm]
\bottomrule
 \end{tabular}}
\end{sidewaystable*}

\begin{sidewaystable*}[htbp]
\centering
\caption{Summary of the strengths and weaknesses of the existing fire and smoke datasets (continued).}
\vspace{1mm}
\label{Table: proscons2}
\resizebox{\textwidth}{!}{
\setlength{\tabcolsep}{7pt}
\begin{tabular}{lllllllll}
\toprule
\textbf{No.}    &\textbf{Dataset}   & \textbf{Pros} & \textbf{Cons}   \\
\midrule 

14     
&FireNet
&\tabitem Provides fire-labeled data, facilitating supervised learning for fire detection.
&\tabitem Comprises only 502 images, which is insufficient for training deep learning models.
\\[1mm]
&&\tabitem Designed specifically for training models with fire detection in real-time.
&\tabitem Data collected from a fixed resolution that limits the model generalizability.
\\[3mm]

\myrowcolour
15       
&AIDER
&\tabitem Sources from multiple platforms, offering diverse perspectives and conditions.
&\tabitem Only 740 of 8,540 images show fire/smoke, possibly affecting model training.
\\[1mm]
\myrowcolour
&&\tabitem Despite being collected from various sources, images are standardized.
&\tabitem Images standardized to $128p\times128p$ which impacts the high-resolution models.
\\[3mm]

16     
&ForestryImage
&\tabitem Captures a broad range of flora and fauna found in forest environments.
&\tabitem Contains only 730 fire-labeled images, which can limit the model's robustness.
\\[1mm]
&&\tabitem Provides images with varying resolutions, suitable for detailed analysis.
&\tabitem Focuses on forest ecosystems, with fire-labeled, limiting fire-specific applications.
\\[3mm]

\myrowcolour
17       
&FIRE
&\tabitem The dataset is mainly structured to facilitate binary classification tasks.
&\tabitem The dataset's small size may lead to overfitting and limit the generalizability.
\\[1mm]
\myrowcolour
&&\tabitem Captured under different brightness levels, smoke density, and flame intensity.
&\tabitem Contains 755 fire and 244 non-fire images, resulting in biased model predictions.
\\[3mm]

18     
&CAIR
&\tabitem Encompasses various fire situations, differing in intensity and luminosity.
&\tabitem Imbalanced between fire and non-fire images may lead to biased model training.
\\[1mm]
&&\tabitem Provides diverse non-fire environments to support training false positives.
&\tabitem Absence of metadata such as location and time limits the dataset’s applicability.
\\[3mm]

\myrowcolour
19       
&FESB MILD
&\tabitem Includes hand-labeled ground truths, supporting semantic segmentation tasks.
&\tabitem Dataset's small size may limit the robustness and generalizability of models.
\\[1mm]
\myrowcolour
&&\tabitem Includes 11 segmentation classes, offering a comprehensive ecological analysis.
&\tabitem Sourced from single-location data, which limits applicability to diverse scenarios.
\\[3mm]

20     
&FiSmo
&\tabitem Integrates multiple datasets, offering a wide range of fire and smoke scenarios.
&\tabitem Comprises various sources, resulting in quality and resolution inconsistencies.
\\[1mm]
&&\tabitem Provides frame-by-frame labels and class categories, enabling deep analysis.
&\tabitem Due to the low resolution for some datasets, detail detection may be affected.
\\[3mm]

\myrowcolour
21       
&Corsican
&\tabitem Includes both visible and near-infrared, enhancing the dataset's applicability.
&\tabitem Includes a limited number of images, possibly insufficient for training DL models.
\\[1mm]
\myrowcolour
&&\tabitem Offers a broad spectrum of scenarios that are useful for robust model training.
&\tabitem Manually extracted ground truth, causing biases or inconsistencies in annotations.
\\[3mm]

22
&FireSense
&\tabitem Encompasses various fire situations, including forest fires and indoor fires.
&\tabitem Limited unique fire scenarios may lead to overfitting and reduce generalizability. 
\\[1mm]
&&\tabitem Includes both fire and smoke samples with labels, offering model versatility.
&\tabitem Contains only 49 videos, which may be insufficient for training robust DL models.
\\[3mm]

\myrowcolour
23       
&BoWFire
&\tabitem Combines pixel color and texture features, enhancing the detection accuracy.
&\tabitem Small size of the dataset may cause overfitting and limit the generalization.
\\[1mm]
\myrowcolour
&&\tabitem Includes fire-like non-fire images, helping reduce false positives in training.
&\tabitem Low image resolution fails to capture sufficient detail for effective model training.
\\[3mm]

24   
&MIVIA
&\tabitem Includes both real-world and web-sourced videos, offering broad datasets.
&\tabitem Lacks diversity in environmental conditions, reducing the dataset's generalizability.
\\[1mm]
&&\tabitem Contains videos with fire-like elements, aiding robust model development.
&\tabitem The slight class imbalance in the dataset could impact the model's performance.
\\[3mm]

\myrowcolour
25       
&VisiFire
&\tabitem Includes diverse range of videos of outdoor fires, forest fires, and indoor fires.
&\tabitem Contains similar scenarios, potentially limiting the generalizability of models.
\\[1mm]
\myrowcolour
&&\tabitem The dataset is specifically designed for analyzing surveillance camera footage.
&\tabitem The dataset's small size may fail to capture real-world fire variability.
\\[3mm]

26   
&FireClips
&\tabitem Provides foundational data for developing binary fire classification algorithms.
&\tabitem Contains only 13 video clips, which restricts its utility for training DL models.
\\[1mm]
&&\tabitem Includes actual footage from rural and active wildfire environments.
&\tabitem The .AVI format may not be compatible with modern video processing tools.
\\[3mm]
\bottomrule
 \end{tabular}}
\end{sidewaystable*}

\clearpage
\section{Experimental Analysis and Discussion}
\label{sec: result}
In this section, we present the experimental results and analysis of fire and smoke datasets to evaluate their effectiveness and robustness in various fire management tasks. The goal is to investigate and assess the performance of CV-based techniques, including classification, segmentation, and detection methods on the fire and smoke datasets. By applying state-of-the-art models and standardized evaluation metrics, we compare their performance across different datasets, highlighting their strengths and limitations. This comparative analysis provides valuable insights into the Applicability of each dataset for specific applications. The subsequent subsections will delve into further detail on evaluation criteria, experimental results, and discussion.

\subsection{Evaluation Criteria}

In this subsection, we outline the evaluation criteria used to assess the performance of the classification, segmentation, and detection algorithms on fire and smoke datasets. To measure various aspects of model effectiveness and ensure the highest confidence in the results, we selected the following 7 metrics:

\begin{itemize}

    \item \textbf{Accuracy}: This metric measures the proportion of correctly classified samples compared to the total number of samples. It provides an overall effectiveness of the model in classification and segmentation tasks. Accuracy can range from $0$ to $1$, with values closer to $1$ indicating better performance. Accuracy formula is as follows:

    \begin{equation} 
    \operatorname{Accuracy} = \frac{\text{TP} + \text{TN}}{\text{TP} + \text{TN} + \text{FP} + \text{FN}}
    \end{equation}
    where $\text{TP}$, $\text{TN}$, $\text{FP}$, and $\text{FN}$ denote the true positives, true negatives, false positives, and false negatives, respectively.

    \item \textbf{Precision}: This metric quantifies the proportion of true positive predictions out of all positive predictions made by the model. It evaluates the model's ability to avoid false positives. Precision values range from $0$ to $1$, with higher values indicating fewer false positives. The formula for precision is given by:

    \begin{equation} 
    \operatorname{Precision} = \frac{\text{TP}}{\text{TP} + \text{FP}} 
    \end{equation}
    where $\text{TP}$ and $\text{FP}$ represent true positives and false positives, respectively.

    \item \textbf{Specificity}: This metric, also known as the True Negative Rate (TNR), measures the proportion of correctly identified negative samples out of all actual negative samples. It evaluates the model's ability to avoid false positives. Specificity ranges from $0$ to $1$, where higher values indicate better performance in correctly identifying negative cases. The specificity formula is as follows:

    \begin{equation}
    \operatorname{Specificity} = \frac{\text{TN}}{\text{TN} + \text{FP}} 
    \end{equation}

    where $\text{TN}$ represents the true negatives, and $\text{FP}$ represents the false positives. Specificity is particularly important in scenarios where minimizing false positives is critical, such as in fire detection and segmentation systems.

    \item \textbf{Recall}: This metric, also known as sensitivity, measures the proportion of true positive predictions out of all actual positive cases. It evaluates the model's ability to detect all relevant instances. The Recall takes values in $[-1, 1]$, where higher values indicate fewer false negatives. The formula for the recall is:

    \begin{equation} 
    \operatorname{Recall} = \frac{\text{TP}}{\text{TP} + \text{FN}} 
    \end{equation}
    where $\text{TP}$ and $\text{FN}$ are true positives and false negatives, respectively.

    \item \textbf{F1-Score}: This metric is the harmonic mean of precision and recall, providing a balanced metric that accounts for both false positives and false negatives. It is particularly useful when the class distribution is imbalanced. The F1-Score ranges from $0$ to $1$, with higher values indicating better performance. The F1-Score formula is:

    \begin{equation} 
    \operatorname{F1-Score} = \frac{2 \cdot \text{Precision} \cdot \text{Recall}}{\text{Precision} + \text{Recall}} 
    \end{equation}
    where precision and recall are as defined above.

    \item \textbf{Intersection over Union (IoU)}: This metric measures the overlap between the predicted and ground truth regions. It is commonly used in detection and segmentation tasks to evaluate how closely two regions are aligned \cite{boroujeni2024ic}. The IoU lies in the range $[0,1]$, where $0$ means weak alignment, and $1$ means perfect alignment with great overlap. IoU for two regions $A$ and $B$ is defined as follows:

    \begin{equation}
    \operatorname{IoU}=\frac{|A \cap B|}{|A \cup B|}.
    \end{equation}
    where $A$ and $B$ are the sets of predicted and ground truth pixels, respectively.

    \item \textbf{Mean Average Precision(mAP)}: This metric evaluates the precision-recall trade-off across multiple IoU thresholds, providing a comprehensive measure of model performance in detection tasks. It is calculated by averaging the Average Precision (AP) values over all classes and IoU thresholds. The formula for mAP is:

    \begin{equation} 
    \operatorname{mAP} = \frac{1}{N} \sum_{i=1}^{N} \text{AP}_i 
    \end{equation}

    where $N$ is the number of classes, and $\text{AP}_i$ is the Average Precision for the $i$-th class, which is computed as the area under the precision-recall curve.

    \textbf{mAP(50)} metric evaluates the mean Average Precision at a fixed IoU threshold of 0.50. It focuses on assessing how well the model detects objects with a moderate level of localization accuracy. A higher mAP(50) indicates that the model is effective at detecting objects with relatively moderate bounding box alignment. On the other hand, \textbf{mAP(50-95)} metric provides a more comprehensive evaluation by averaging the mAP over multiple IoU thresholds ranging from $0.50$ to $0.95$, in increments of 0.05. This metric requires both accurate object detection and precise bounding box localization. 
    
\end{itemize}

\subsection{Results and discussion}

In this section, we evaluate the performance of classification, segmentation, and detection algorithms on the fire and smoke datasets. We evaluate the efficiency of the examined models through various metrics to provide a comprehensive analysis of their effectiveness, robustness, generalizability, as well as their limitations.

\begin{figure}[H]
    \centering
    \centerline{\includegraphics[width=0.9\textwidth]
    {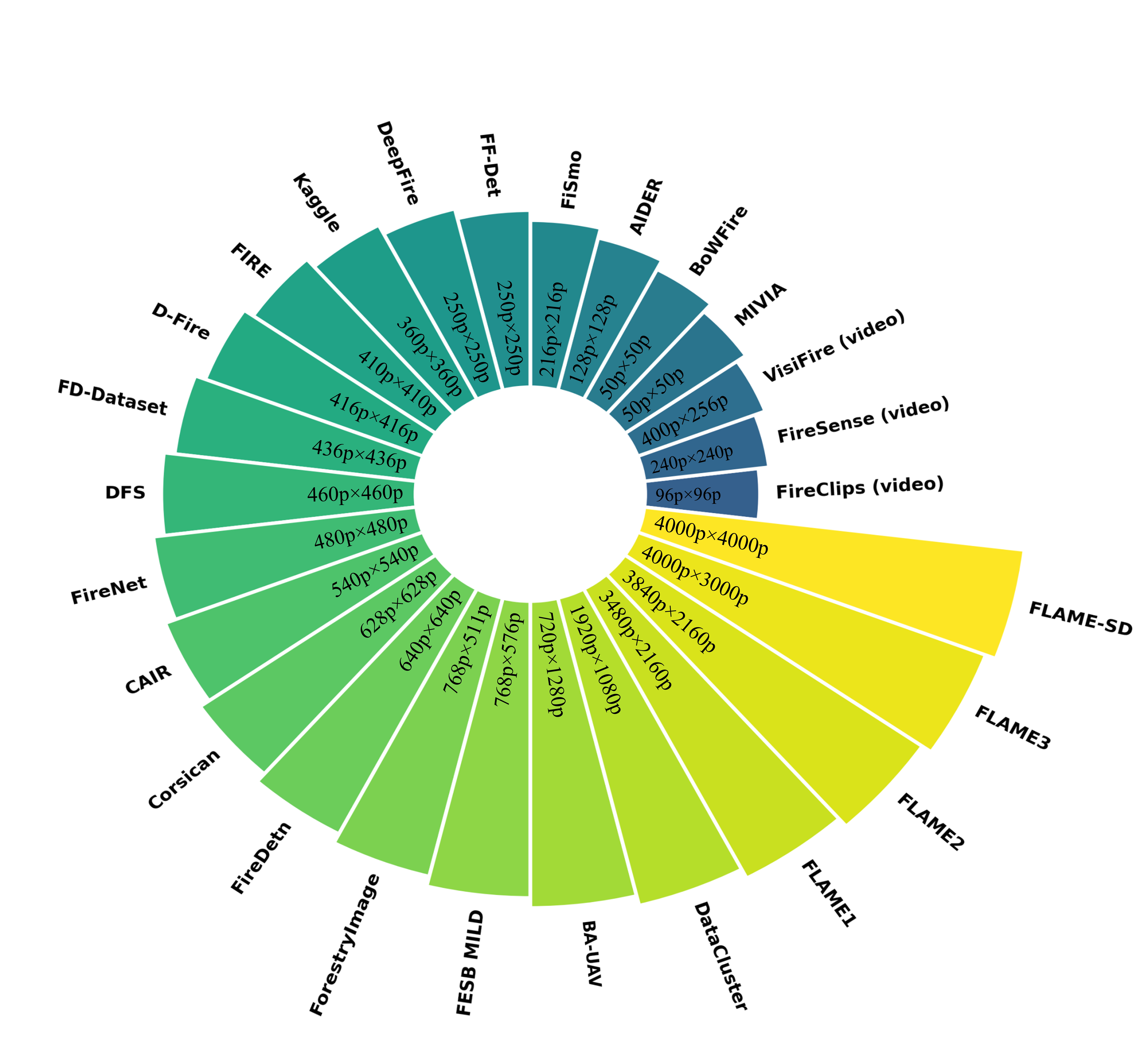}}
    \caption{Ranking of the fire and smoke datasets based on image resolution and quality. The bar heights represent the relative resolution and quality of each dataset. Higher bars indicate superior resolution, while lower bars represent lower resolution.}
    \label{fig: graph}
\end{figure}

Figure \ref{fig: graph} ranks various fire and smoke datasets based on their image resolution and quality. The datasets are represented in a circular arrangement, with their names around the plot. The height of each bar corresponds to the relative quality and resolution of the dataset, where higher bars indicate datasets of higher quality. From \ref{fig: graph}, it is clear that datasets like FLAME-SD, FLAME3, and FLAME2 stand out with the highest bar heights, signifying superior resolution and image clarity. This characteristic makes them more suitable for tasks that require detailed fine-grained. In contrast, datasets like FireClips, VisiFire, and FireSense have shorter bars, indicating relatively lower resolution and quality. The dynamic coloring further emphasizes the differences, transitioning from cooler blues for lower-quality datasets to vibrant yellows for higher-quality ones.

\begin{figure}[H]
    \centering
    \centerline{\includegraphics[width=0.9\textwidth]
    {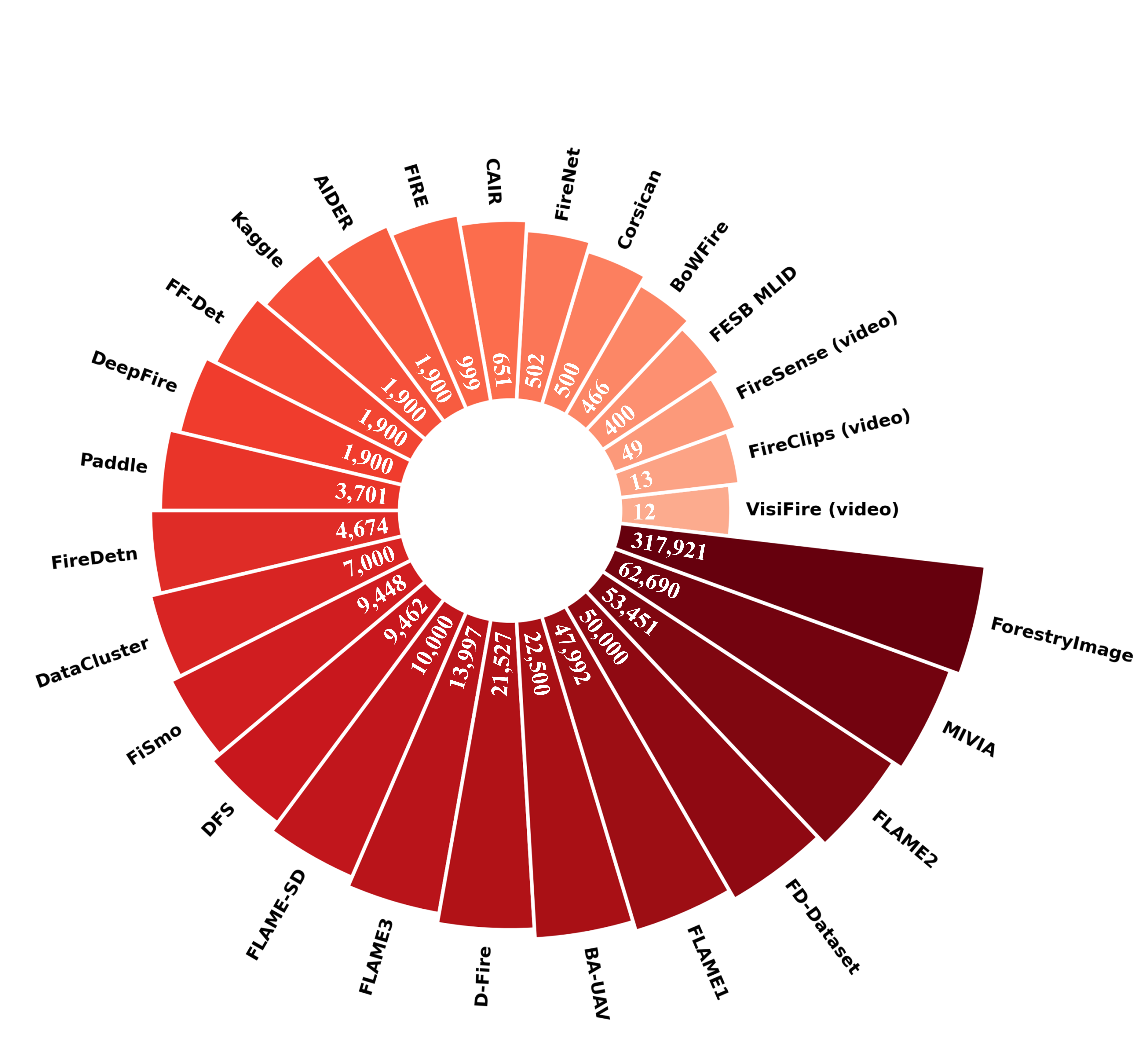}}
    \caption{Ranking of the fire and smoke datasets according to the number of images per dataset. The bar heights represent the relative sizes of the datasets, with taller bars indicating larger datasets and shorter bars representing smaller ones.}
    \label{fig: sizegraph}
\end{figure}

Figure \ref{fig: sizegraph} ranks various fire and smoke datasets according to the number of images per dataset. The datasets are represented in a circular arrangement, with their names around the plot. The height of each bar corresponds to the relative sizes of the dataset, where taller bars indicate larger datasets and shorter bars represent smaller ones. From \ref{fig: sizegraph}, it is obvious that datasets like ForestryImage, MIVIA, and FLAME2 stand out with the highest bar heights, signifying large datasets. This characteristic makes them more suitable for tasks that require a substantial amount of data, such as training deep learning models for fire and smoke detection or segmentation. Conversely, datasets with shorter bars, such as VisiFire and FireClips, may be more appropriate for tasks requiring lightweight models. The dynamic coloring further emphasizes the differences, transitioning from cooler reds for smaller datasets to vibrant reds for larger ones.

Table \ref{Table: Classification} presents an in-depth analysis of fire and smoke datasets for the classification task. The experimental results are obtained by training the ResNet-50 algorithm on the training datasets (first column) and then evaluating the model's performance on the testing datasets (first row). The performance of each dataset is assessed using accuracy, precision, recall, and F1-Score metrics to provide a comprehensive assessment. For convenience in comparison, we also ranked the datasets based on the accuracy metric from 1 (best) to 11 (worst).

\begin{table}[H]
\centering
\caption{The comparative analysis of fire and smoke datasets for classification tasks: ResNet-50 performance across Training (left) and Testing (top) datasets.}
\vspace{1mm}
\label{Table: Classification}
\resizebox{\textwidth}{!}{
\setlength{\tabcolsep}{7pt}
\begin{tabular}{lllllllllllllllllll}
\toprule

\textbf{Training}      &\textbf{Eval}      &&&&&  \textbf{Testing}  & \textbf{Datasets} \\

\cmidrule(l){3-13}

\textbf{Dataset}   & \textbf{Metric}    &  \textbf{Flame1} & \textbf{Flame2} &\textbf{Flame3}    & \textbf{DeepFire}  & \textbf{D-Cluster}   & \textbf{FF-Det}   & \textbf{FIRE}   & \textbf{Kaggle}   & \textbf{BoWFire}   & \textbf{D-Fire}    & \textbf{FiSmo}   \\
\midrule 

\textbf{Flame1}     
& Accuracy 
& \textbf{98.49\%} 
& 74.70\%
& 84.28\%
& 98.42\%
& 93.94\%
& 66.18\%
& 53.74\%
& 48.66\%
& 23.95\%
& 95.90\%
& 92.48\%
& 
\\[1mm]
& Precision 
& \textbf{100\%}
& 75\%
& 84\%
& 100\%
& 100\%
& 49\%
& 23\%
& 68\%
& 100\%
& 54\%
& 100\%
&\\[1mm]
& Recall 
& \textbf{98\% }
& 100\%
& 100\%
& 98\%
& 94\%
& 94\%
& 92\%
& 96\%
& 92\%
& 95\%
& 96\%
&\\[1mm]
& F1-Score 
& \textbf{99\%}
& 86\%
& 91\%
& 99\%
& 97\%
& 65\%
& 37\%
& 79\%
& 96\%
& 69\%
& 98\%
& \\[1mm]
& Rank
& \textbf{(1)}
& (7)
& (6)
& (2)
& (4)
& (8)
& (9)
& (10)
& (11)
& (3)
& (5)
& \\[2.5mm]

\myrowcolour
\textbf{Flame2}     
& Accuracy
& 40.39\%
& \textbf{98.54\%} 
& 41.49\%
& 48.95\%
& 52.53\%
& 51.32\%
& 23.25\%
& 44.12\%
& 49.56\%
& 50.81\%
& 58.10\%
&\\[1mm]

\myrowcolour
& Precision 
& 16\%
& \textbf{100\%}
& 69\%
& 100\%
& 100\%
& 51\%
& 20\%
& 61\%
& 100\%
& 55\%
& 100\%
&\\[1mm]
\myrowcolour
& Recall 
& 40\%
& \textbf{99\%}
& 31\%
& 49\%
& 53\%
& 46\%
& 50\%
& 48\%
& 50\%
& 41\%
& 58\%
&\\[1mm]
\myrowcolour
& F1-Score  
& 23\%
& \textbf{99\%}
& 43\%
& 66\%
& 69\%
& 49\%
& 30\%
& 54\%
& 66\%
& 47\%
& 74\%
& \\[1mm]
\myrowcolour
& Rank
& (10)
& \textbf{(1)}
& (9)
& (7)
& (3)
& (4)
& (11)
& (8)
& (6)
& (5)
& (2)
& \\[2.5mm]

\textbf{Flame3}     
& Accuracy
& 57.33\% 
& 69.80\%
& \textbf{98.19\%}
& 42.63\%
& 29.29\%
& 65.79\%
& 61.48\%
& 72.06\%
& 56.64\%
& 47.75\%
& 19.40\%
&\\[1mm]
& Precision 
& 51\%
& 80\%
& \textbf{100\%}
& 100\%
& 100\%
& 66\%
& 24\%
& 80\%
& 100\%
& 51\%
& 100\%
&\\[1mm]
& Recall 
& 52\%
& 79\%
& \textbf{99\%}
& 43\%
& 29\%
& 64\%
& 27\%
& 78\%
& 57\%
& 65\%
& 19\%
&\\[1mm]
& F1-Score  
& 57\%
& 80\%
& \textbf{99\%}
& 60\%
& 45\%
& 65\%
& 25\%
& 79\%
& 72\%
& 57\%
& 33\%
& \\[1mm]
& Rank
& (6)
& (3)
& \textbf{(1)}
& (9)
& (10)
& (4)
& (5)
& (2)
& (7)
& (8)
& (11)
& \\[2.5mm]

\myrowcolour
\textbf{DeepFire}     
& Accuracy
& 45.36\%
& 68.21\%
& 84.55\%
& \textbf{96.54\%}
& 35.35\%
& 95.84\%
& 95.81\%
& 96.53\%
& 25.22\%
& 40.90\%
& 04.10\%
&\\[1mm]
\myrowcolour
& Precision 
& 67\%
& 77\%
& 88\%
& \textbf{100\%}
& 100\%
& 97\%
& 87\%
& 98\%
& 100\%
& 46\%
& 100\%
&\\[1mm]
\myrowcolour
& Recall 
& 45\% 
& 82\%
& 95\%
& \textbf{96\%}
& 35\%
& 96\%
& 98\%
& 100\%
& 25\%
& 68\%
& 40\%
&\\[1mm]
\myrowcolour
& F1-Score  
& 35\%
& 79\%
& 91\%
& \textbf{94\%}
& 52\%
& 97\%
& 92\%
& 99\%
& 40\%
& 55\%
& 90\%
& \\[1mm]
\myrowcolour
& Rank
& (7)
& (6)
& (5)
& \textbf{(1)}
& (9)
& (3)
& (4)
& (2)
& (10)
& (8)
& (11)
& \\[2.5mm]

\textbf{D-Cluster}     
& Accuracy 
& 59.52\% 
& 51.19\%
& 60.22\%
& 50.53\%
& \textbf{94.21\%}
& 53.87\%
& 73.29\%
& 23.54\%
& 33.55\%
& 67.90\%
& 71.21\%
&\\[1mm]
& Precision 
& 62\%
& 100\%
& 60\%
& 50\%
& \textbf{100\%}
& 53\%
& 67\%
& 32\%
& 33\%
& 63\%
& 73\%
&\\[1mm]
& Recall 
& 40\%
& 49\%
& 65\%
& 53\%
& \textbf{94\%}
& 46\%
& 50\%
& 48\%
& 50\%
& 41\%
& 58\%
&\\[1mm]
& F1-Score  
& 100\%
& 76\%
& 68\%
& 67\%
& \textbf{97\%}
& 69\%
& 84\%
& 38\%
& 50\%
& 87\%
& 83\%
& \\[1mm]
& Rank
& (6)
& (8)
& (5)
& (9)
& \textbf{(1)}
& (7)
& (2)
& (11)
& (10)
& (4)
& (3)
& \\[2.5mm]

\myrowcolour
\textbf{FF-Det}     
& Accuracy
& 58.96\% 
& 42.18\%
& 63.69\%
& 45.53\%
& 23.23\%
& \textbf{99.30\%}
& 94.41\%
& 98.48\%
& 27.43\%
& 39.46
& 03.00\%
&\\[1mm]
\myrowcolour
& Precision 
& 43\%
& 63\%
& 80\%
& 100\%
& 100\%
& \textbf{100\%}
& 89\%
& 100\%
& 100\%
& 45\%
& 100\%
&\\[1mm]
\myrowcolour
& Recall 
& 59\% 
& 56\%
& 75\%
& 46\%
& 23\%
& \textbf{94\%}
& 88\%
& 97\%
& 27\%
& 56\%
& 13\%
&\\[1mm]
\myrowcolour
& F1-Score  
& 45\%
& 59\%
& 78\%
& 63\%
& 38\%
& \textbf{95\%}
& 88\%
& 98\%
& 43\%
& 50\%
& 16\%
& \\[1mm]
\myrowcolour
& Rank
& (5)
& (7)
& (4)
& (6)
& (10)
& \textbf{(1)}
& (3)
& (2)
& (9)
& (8)
& (11)
& \\[2.5mm]

\textbf{Fire}     
& Accuracy 
& 46.00\%
& 74.30\%
& 68.33\%
& 91.23\%
& 88.43\%
& 87.89\%
& \textbf{92.29\%}
& 82.22\%
& 81.09\%
& 62.96\%
& 91.40\%
&\\[1mm]
& Precision 
& 97\%
& 74\%
& 88\%
& 89\%
& 100\%
& 85\%
& \textbf{77\%}
& 69\%
& 100\%
& 75\%
& 100\%
&\\[1mm]
& Recall 
& 70\%
& 78\%
& 75\%
& 94\%
& 100\%
& 90\%
& \textbf{96\%}
& 82\%
& 82\%
& 44\%
& 91\%
&\\[1mm]
& F1-Score  
& 85\%
& 80\%
& 85\%
& 91\%
& 100\%
& 87\%
& \textbf{85\%}
& 75\%
& 90\%
& 56\%
& 95\%
& \\[1mm]
& Rank
& (11)
& (8)
& (10)
& (3)
& (4)
& (5)
& \textbf{(1)}
& (6)
& (7)
& (9)
& (2)
& \\[2.5mm]

\myrowcolour
\textbf{Kaggle}     
& Accuracy 
& 40.39\% 
& 67.50\%
& 83.20\%
& 48.42\%
& 35.35\%
& 97.89
& 96.21\%
& \textbf{99.56\%}
& 30.97\%
& 41.08\%
& 05.50\%
&\\[1mm]
\myrowcolour
& Precision 
& 16\%
& 73\%
& 84\%
& 100\%
& 100\%
& 97\%
& 92\%
& \textbf{100\%}
& 100\%
& 46\%
& 44\%
&\\[1mm]
\myrowcolour
& Recall 
& 40\%
& 90\%
& 99\%
& 48\%
& 35\%
& 99\%
& 93\%
& \textbf{96\%}
& 31\%
& 68\%
& 06\%
&\\[1mm]
\myrowcolour
& F1-Score  
& 23\%
& 81\%
& 91\%
& 65\%
& 52\%
& 98\%
& 92\%
& \textbf{97\%}
& 47\%
& 55\%
& 10\%
& \\[1mm]
\myrowcolour
& Rank
& (8)
& (5)
& (4)
& (6)
& (9)
& (2)
& (3)
& \textbf{(1)}
& (10)
& (7)
& (11)
& \\[2.5mm]

\textbf{BoWFire}     
& Accuracy 
& 42.11\% 
& 74.37\%
& 78.88\%
& 62.91\%
& 51.49\%
& 55.73\%
& 37.13\%
& 72.05\%
& \textbf{93.70\%}
& 62.96\%
& 64.11\%
&\\[1mm]
& Precision 
& 42\%
& 74\%
& 85\%
& 57\%
& 80\%
& 53\%
& 28\%
& 100\%
& \textbf{100\%}
& 81\%
& 85\%
&\\[1mm]
& Recall 
& 85\%
& 100\%
& 100\%
& 99\%
& 90\%
& 83\%
& 78\%
& 13\%
& \textbf{90\%}
& 19\%
& 17\%
&\\[1mm]
& F1-Score  
& 59\%
& 85\%
& 93\%
& 72\%
& 78\%
& 69\%
& 44\%
& 24\%
& \textbf{94\%}
& 38\%
& 28\%
& \\[1mm]
& Rank
& (10)
& (3)
& (2)
& (7)
& (9)
& (8)
& (11)
& (5)
& \textbf{(1)}
& (6)
& (5)
& \\[2.5mm]

\myrowcolour
\textbf{D-Fire}     
& Accuracy 
& 40.15\%
& 69.16\%
& 84.01\%
& 59.47\%
& 82.83\%
& 29.21\%
& 11.38\%
& 32.35\%
& 71.24\%
& \textbf{98.03\%}
& 90.10\%
& \\[1mm]
\myrowcolour
& Precision 
& 33\%
& 78\%
& 87\%
& 100\%
& 100\%
& 35\%
& 08\%
& 50\%
& 100\%
& \textbf{100\%}
& 100\%
&\\[1mm]
\myrowcolour
& Recall 
& 40\% 
& 83\%
& 95\%
& 59\%
& 83\%
& 47\%
& 23\%
& 41\%
& 71\%
& \textbf{93\%}
& 90\%
&\\[1mm]
\myrowcolour
& F1-Score  
& 23\%
& 80\%
& 91\%
& 75\%
& 91\%
& 40\%
& 11\%
& 45\%
& 83\%
& \textbf{94\%}
& 95\%
& \\[1mm]
\myrowcolour
& Rank
& (8)
& (6)
& (3)
& (7)
& (4)
& (10)
& (11)
& (9)
& (5)
& \textbf{(1)}
& (2)
& \\[2.5mm]

\textbf{FiSmo}     
& Accuracy 
& 59.21\%
& 32.57\%
& 53.76\%
& 17.80\%
& 43.43\%
& 16.29\%
& 18.18\%
& 15.01\%
& 42.97\%
& 38.88\%
& \textbf{94.36\%}
& \\[1mm]
& Precision 
& 94\%
& 40\%
& 90\%
& 20\%
& 80\%
& 20\%
& 20\%
& 17\%
& 100\%
& 45\%
& \textbf{94\%}
&\\[1mm]
& Recall 
& 50\% 
& 43\%
& 93\%
& 17\%
& 43\%
& 10\%
& 20\%
& 17\%
& 43\%
& 45\%
& \textbf{93\%}
&\\[1mm]
& F1-Score  
& 37\%
& 17\%
& 61\%
& 20\%
& 61\%
& 10\%
& 30\%
& 25\%
& 60\%
& 53\%
& \textbf{93\%}
& \\[1mm]
& Rank
& (2)
& (7)
& (3)
& (9)
& (4)
& (10)
& (8)
& (11)
& (5)
& (6)
& \textbf{(1)}
& \\[2.5mm]

\midrule  
\textbf{Ranking}
& \textbf{Average}
& \textbf{6.73}
& \textbf{5.55}
& \textbf{4.73}
& \textbf{6.00}
& \textbf{6.09}
& \textbf{5.64}
& \textbf{6.18}
& \textbf{6.09}
& \textbf{7.36}
& \textbf{5.91}
& \textbf{5.82}
& \\[2.5mm]

\bottomrule
 \end{tabular}}
\end{table}

From Table \ref{Table: Classification}, it is evident that the FLAME dataset family (FLAME1, FLAME2, and FLAME3) plays a crucial role in understanding dataset adaptability across different fire and smoke classification tasks. FLAME1 achieves the highest performance on its own test set, with an accuracy of 98.49\% and an F1-score of 99\%, ranking first among the training datasets. However, its generalization capability varies across different test datasets. While it maintains high accuracy on DeepFire (98.42\%), D-Cluster (93.94\%), and D-Fire (90.96\%), it struggles on datasets like BoWFire (23.95\%) and Kaggle (48.66\%). It can be concluded that FLAME1 is well-structured for its training environment but lacks robustness when applied to datasets with significantly different distributions. Similarly, FLAME2 exhibits strong performance on its own test set (Accuracy: 98.19\%, Precision: 100\%), while it underperforms on datasets such as FLAME1 (40.39\%) and FIRE (23.25\%). The sharp contrast in its performance across different datasets suggests that FLAME2 might be biased toward specific fire characteristics, limiting its ability to generalize effectively. By the same observation from the table, it is clear that the FLAME3 obtains the highest accuracy (98.19\%) on its own test set, as well as the lowest accuracy on FiSmo (19.40\%) and D-Cluster (29.29\%) datasets. Despite these two lower-performing datasets, FLAME3 maintains an accuracy of over 50\% on all other datasets, indicating strong generalizability compared to many other training datasets. Its ability to generalize across most datasets, except for the two lowest-ranked ones, suggests that FLAME3 contains diverse enough features to support cross-dataset adaptability while still facing challenges in highly varied environments like D-Cluster and FiSmo.

DeepFire demonstrates fair generalization on external datasets such as FF-Det (95.84\%), Kaggle (96.53\%), and FIRE (95.81\%) datasets. However, it performs significantly poorly on the FiSmo and BoWFire datasets, with an accuracy of 4.10\% and 25.22\%, respectively. This indicates that DeepFire captures useful features for classification but may require additional data diversity to improve its robustness further. On the other hand, D-Cluster demonstrates a balanced performance across multiple test datasets, achieving high accuracy on its own set (94.21\%) and strong results when tested on datasets such as FIRE (73.29\%) and FiSmo (71.21\%). This dataset's consistent performance across different test environments suggests it may contain a diverse range of fire and smoke samples, making it a more generalizable dataset for training deep learning models.

FF-Det achieves an outstanding accuracy of 99.30\% when tested on its own dataset, along with strong performance on Kaggle (98.48\%) and FIRE (94.41\%). Its consistently high accuracy and F1-score indicate that FF-Det provides a diverse and representative dataset for training models that can generalize well to various fire and smoke scenarios. This makes it an ideal dataset for training models with robust cross-dataset adaptability. In contrast, the FIRE dataset, while performing well on its own dataset (92.29\% accuracy), exhibits a moderate generalization capability. It performs very well on FiSmo (91.40\%) and DeepFire (91.23\%), indicating that it shares some feature similarities with these datasets. However, its performance drops slightly on FLAME1 (46.00\%) and FLAME3 (68.33\%), suggesting that the dataset lacks variability in certain fire conditions. Last but not least, the FIRE dataset demonstrates notable generalization performance across all test datasets, maintaining an accuracy of almost 60\% on all but one dataset.

Kaggle maintains high accuracy on its own dataset (99.56\%) and generalizes well to FF-Det (97.89\%), but struggles on FiSmo (05.50\%) and BoWFire (30.70\%). These results suggest that Kaggle contains a diverse range of fire scenarios but might not be well-suited for extremely challenging environments like FiSmo. Its high self-test performance and strong results on certain external datasets indicate that it is a solid dataset for training but requires additional variations to improve adaptability to highly diverse fire conditions. BoWFire exhibits one of the weakest generalization performances across all datasets. It achieves 93.70\% accuracy on its own dataset, which is among the lowest self-test scores. Additionally, its performance drops significantly on FIRE (37.13\%) and FLAME1 (42.11\%), suggesting that it contains highly variable or complex fire and smoke conditions that make classification more difficult. Its consistently low ranking highlights the need for data augmentation or additional pre-processing to improve model learning.

D-Fire generally ranks among the highest-performing datasets, both in self-testing (accuracy: 98.03\%) and across external datasets. The model struggles when tested on FIRE, with performance dropping to 11.38\% when trained on FIRE and 29.21\% when trained on FF-Det. These results suggest that D-Fire presents significant challenges due to the complexity or high intra-class variability of its images. It serves as a challenging benchmark for fire detection models but requires additional improvements to be effective for training purposes. Meanwhile, FiSmo achieves a fair accuracy (94.36\%) when tested on its own set; and also performs well on FLAME1 (59.21\%) and FLAME3 (53.76\%) datasets. However, its performance is notably lower on Kaggle (15.01\%) and FF-Det (16.29\%), indicating that it may not fully cover certain fire and smoke conditions. An interesting point about FiSmo is that, despite the trained model on this dataset obtaining a low accuracy of 50\% when tested on all other datasets, it is among the top-performing datasets in terms of generalization, demonstrating high accuracy across most datasets when tested on this dataset.

Table \ref{Table: Segmentation} provides a comparative analysis of fire and smoke datasets for the segmentation task. The experimental results are achieved by training the DeepLab-V3 algorithm on the training datasets (first column) and then assessing the model's performance on the testing datasets (first row). The performance of each dataset is evaluated using accuracy, specificity, F1-Score, and IoU metrics to provide an in-depth assessment. For convenience in comparison, we also ranked the datasets based on the IoU metric from 1 (best) to 4 (worst). Additionally, the table includes detailed information about the training, validation, and test subsets of each dataset used throughout the experimental process.

From Table \ref{Table: Segmentation}, it is evident that the model is initially trained on the FLAME1 dataset. It achieves satisfactory performance on its own test set (IoU of 88.22\%, accuracy of 99.84\%, and F1-Score of 92.91\%). However, the model's generalizability significantly decreases on the other datasets, particularly on the FESB MILD dataset, where the IoU and F1-Score are only 19.54\% and 6.02\%, respectively. This suggests that the FLAME1 dataset requires more diverse training data to improve the model's adaptability and generalizability across different scenarios. Meanwhile, the BA-UAV dataset shows similar performance trends when evaluated on its own test set, achieving an IoU of 53.12\%, accuracy of 75.97\%, F1-Score of 71.84\%, and 85.21\% in specificity. Although its performance on the other datasets is better compared to the FLAME1 dataset, it still indicates a lack of the required robustness for effective generalization to datasets with diverse features.

By the same observation from Table \ref{Table: Segmentation}, it is clear that the BoWFire dataset not only performs very well on its own test set (the highest IoU of 89.20\% and accuracy of 99.89\%) but also comparatively achieves better results when tested on other datasets. In particular, the model obtains an accuracy of 99.41\% on the FLAME1 dataset, indicating a certain degree of feature overlap or similarity between these datasets. These findings highlight the dataset's potential for training models with improved generalization capabilities across diverse environments. In the meantime, the FESB MILD dataset serves as a fair resource for model training and testing despite being the smallest in size. The model achieves moderate performance (IoU of 75.30\% and F1-Score of 81.82\%), which shows that the dataset is internally consistent. However, its generalization to other datasets is notably poor, as evidenced by low accuracy and metric scores on datasets such as FLAME1 and BoWFire. These outcomes highlight the limitations of FESB MILD, such as its limited size and specific fire and smoke patterns, which may not be well-aligned with other datasets.

\begin{table}[H]
\centering
\caption{The comparative analysis of wildfire datasets for segmentation tasks: DeepLab-V3 performance across Training (left) and Testing (top) datasets.}
\vspace{1mm}
\label{Table: Segmentation}
\resizebox{\textwidth}{!}{
\setlength{\tabcolsep}{7pt}
\begin{tabular}{llllllllllll}
\toprule

\textbf{Training}      &\textbf{Evaluation}     &&  \textbf{Testing}  & \textbf{Datasets}   && 
 \hspace{7mm} \textbf{Training} &\textbf{Description}\\

\cmidrule(l){3-6}
\cmidrule(l){7-8}
\textbf{Dataset}   & \textbf{Metric}    &  \textbf{Flame1} & \textbf{BA-UAV} & \textbf{BoWFire}  & \textbf{FESB MILD}  &\textbf{Sub-data Name}   &\textbf{Sub-data Size} \\

\midrule 

\textbf{Flame1}     
& IoU 
& \textbf{88.22\%}
& 26.43\%
& 51.28\% 
& 19.54\%
& Total Set
& 2,003
& \\[1mm]
& F1-Score 
& \textbf{92.91\%}
& 24.53\%
& 55.17\% 
& 06.20\%
& Training Set
& 1,200
&\\[1mm]
& Accuracy
& \textbf{99.84\%}
& 57.56\%
& 94.79\% 
& 32.89\%
& Validation Set
& 400
&\\[1mm]
& Specificity
& \textbf{99.96\%}
& 60.40\%
& 96.18\% 
& 45.46\%
& Test Set
& 400
& \\[1mm]
& Rank
& \textbf{(1)}
& (3)
& (2)
& (4)
& Out of Set
& 3
& \\[2mm]

\myrowcolour
\textbf{BA-UAV}     
& IoU
& 37.23\%
& \textbf{53.12\%}
& 41.30\% 
& 39.84\%
& Total Set
& 46
&\\[1mm]

\myrowcolour
& F1-Score 
& 40.81\%
& \textbf{71.84\%}
& 54.33\% 
& 41.84\%
& Training Set
& 25
&\\[1mm]
\myrowcolour
& Accuracy 
& 66.17\%
& \textbf{75.97\%}
& 82.98\% 
& 72.30\%
& Validation Set
& 10
&\\[1mm]
\myrowcolour
& Specificity 
& 71.55\%
& \textbf{85.21\%}
& 92.53\% 
& 88.15\%
& Test Set
& 10
& \\[1mm]
\myrowcolour
& Rank
& (4)
& \textbf{(1)}
& (2)
& (3)
& Out of Set
& 1
& \\[2mm]

\textbf{BoWFire}     
& IoU 
& 57.20\%
& 40.23\%
& \textbf{89.20\%} 
& 82.39\%
& Total Set
& 226
&\\[1mm]

& F1-Score 
& 55.33\%
& 39.56\%
& \textbf{91.97\%} 
& 87.77\%
& Training Set
& 135
&\\[1mm]

& Accuracy 
& 99.41\%
& 60.63\%
& \textbf{99.89\%} 
& 87.45\% 
& Validation Set
& 45
&\\[1mm]

& Specificity 
& 96.60\%
& 93.15\%
& \textbf{99.95\%} 
& 89.38\%
& Test Set
& 45
& \\[1mm]
& Rank
& (3)
& (4)
& \textbf{(1)}
& (2)
& Out of Set
& 1
& \\[2mm]

\myrowcolour
\textbf{FESB MILD}     
& IoU
& 37.65\%
& 37.36\%
& 50.51\%
& \textbf{75.30\%} 
& Total Set
& 199
&\\[1mm]
\myrowcolour
& F1-Score 
& 38.39\%
& 37.52\%
& 48.72\%
& \textbf{81.82\%} 
& Training Set
& 119
&\\[1mm]
\myrowcolour
& Accuracy 
& 36.06\%
& 60.05\%
& 95.12\%
& \textbf{100\%} 
& Validation Set
& 40
&\\[1mm]
\myrowcolour
& Specificity   
& 85.44\%
& 83.12\%
& 93.56\%
& \textbf{99.80\%} 
& Test Set
& 40
& \\[1mm]
\myrowcolour
& Rank
& (3)
& (4)
& (2)
& \textbf{(1)}
& Out of Set
& 0
& \\[2mm]

\bottomrule
 \end{tabular}}
\end{table}

Table \ref{Table: Detection} presents an in-depth analysis of fire and smoke datasets for the detection task. The experimental results are obtained by training the YoloV8 algorithm on the training datasets (first column) and then evaluating the model's performance on the testing datasets (first row). Each dataset is evaluated on metrics including precision, recall, F1-Score, mAP(50), and mAP(50-95). For convenience in comparison, we also ranked the datasets based on the mAP metric from 1 (best) to 5 (worst). Additionally, the table includes detailed information about the training, validation, and test subsets of each dataset used throughout the experimental process.

From Table \ref{Table: Detection}, it is obvious that the generalization D-Cluster dataset is substantially low when tested on other datasets, with F1-Scores and mAP values falling below 50\% in most cases. This highlights that D-Cluster may include limited common features with other datasets or a limited domain representation. Meanwhile, FireDetn shows relatively fair performance and generalization across the other datasets. It achieves the highest precision (95.90\%) and strong mAP(50) of 76.55\% when tested on itself. However, the mAP(50-95) results are lower than 50\%, which means they have limitations in generalization. Generally, the FireDetn dataset performs relatively better on D-Fire and FLAME3 compared to D-Cluster and FireNet datasets. D-Fire achieves a high precision of 93.20\% and mAP(50) of 50.44\% on its own test set but demonstrates poor performance when tested on other datasets. The results prove that the dataset is effective for training within its scope but lacks transferability to other domains.

On the other hand, FireNet and FLAME3 datasets typically perform better in terms of cross-dataset evaluations compared to D-Cluster and D-Fire. The FireNet dataset achieves a fair mAP(50-95) between 15\% and 55\% on other datasets, while the FLAME3 dataset demonstrates superior performance, with mAP(50-95) ranging from 40\% to 65\% when tested across different datasets. Therefore, we can conclude that the FLAME3 dataset offers better generalization and transferability which makes it more suitable for different scenarios. In conclusion, FLAME3 and FireNet datasets are most appropriate for generalizable wildfire detection tasks due to their consistent performance as well as their data diversities. Conversely, D-Cluster and D-Fire may require augmentation or integration with other datasets to improve their generalization capabilities.

\begin{table}[H]
\centering
\caption{The comparative analysis of wildfire datasets for detection tasks: YoloV8 performance across Training (left) and Testing (top) datasets.}
\vspace{1mm}
\label{Table: Detection}
\resizebox{\textwidth}{!}{
\setlength{\tabcolsep}{7pt}
\begin{tabular}{llllllllllll}
\toprule

\textbf{Training}      &\textbf{Evaluation}     && \hspace{2mm}  \textbf{Testing}  &  \textbf{Datasets}   &&& 
 \hspace{7mm} \textbf{Training} &\textbf{Description}\\

\cmidrule(l){3-7}
\cmidrule(l){8-9}
\textbf{Dataset}   & \textbf{Metric}    &  \textbf{D-Cluster} & \textbf{FireDetn} & \textbf{D-Fire}  & \textbf{FireNet}  & \textbf{Flame3}   &\textbf{Sub-data Name}   &\textbf{Sub-data Size} \\

\midrule 

\textbf{D-Cluster}     
& Precision 
& \textbf{48.60\%}
& 45.91\%
& 13.40\%
& 62.53\%
& 64.22\%
& Total Set
& 7,000
& \\[1mm]
& F1-Score 
& \textbf{50.00\%}
& 26.35\%
& 05.78\%
& 26.36\%
& 39.04\%
& Training Set
& 4,900
&\\[1mm]
& Recall
& \textbf{45.55\%}
& 18.40\%
& 04.51\%
& 16.77\%
& 35.71\%
& Validation Set
& 1,400
&\\[1mm]
& mAP(50)
& \textbf{50.35\%}
& 29.60\%
& 08.22\%
& 38.55\%
& 39.45\%
& Test Set
& 700
& \\[1mm]
& mAP(50-95)
& \textbf{28.90\%}
& 14.66\%
& 03.95\%
& 11.22\%
& 21.44\%
& Out of Set
& 0
& \\[1mm]
& Rank
& \textbf{(1)}
& (4)
& (5)
& (3)
& (2)
& 
& 
& \\[2mm]

\myrowcolour
\textbf{FireDetn}     
& Precision
& 75.93\%
& \textbf{95.90\%}
& 35.11\%
& 77.83\%
& 78.24\%
& Total Set
& 4,674
&\\[1mm]
\myrowcolour
& F1-Score 
& 52.82\%
& \textbf{69.40\%}
& 16.35\%
& 58.44\%
& 55.38\%
& Training Set
& 3,270
&\\[1mm]
\myrowcolour
& Recall 
& 36.40\%
& \textbf{56.65\%}
& 10.63\%
& 46.70\%
& 49.33\%
& Validation Set
& 900
&\\[1mm]
\myrowcolour
& mAP(50)   
& 61.83\%
& \textbf{76.55\%}
& 22.83\%
& 63.63\%
& 58.83\%
& Test Set
& 450
& \\[1mm]
\myrowcolour
& mAP(50-95)
& 41.61\%
& \textbf{49.54\%}
& 12.21\%
& 32.65\%
& 41.24\%
& Out of Set
& 54
& \\[1mm]
\myrowcolour
& Rank
& (3)
& \textbf{(1)}
& (5)
& (2)
& (4)
& 
& 
& \\[2mm]

\textbf{D-Fire}     
& Precision
& 47.31\%
& 58.77\%
& \textbf{93.20\%}
& 50.00\%
& 68.44\%
& Total Set
& 2,1527
\\[1mm]
& F1-Score 
& 23.71\%
& 15.55\%
& \textbf{50.97\%}
& 11.83\%
& 25.91\%
& Training Set
& 1,300
&\\[1mm]
& Recall 
& 13.69\%
& 08.98\%
& \textbf{54.22\%}
& 06.75\%
& 15.55\%
& Validation Set
& 4,300
&\\[1mm]
& mAP(50)   
& 46.00\%
& 32.73\%
& \textbf{50.44\%}
& 28.98\%
& 25.18\%
& Test Set
& 4,300
& \\[1mm]
& mAP(50-95)
& 28.72\%
& 17.24\%
& \textbf{30.55\%}
& 06.11\%
& 30.40\%
& Out of Set
& 73
& \\[1mm]
& Rank
& (2)
& (3)
& \textbf{(1)}
& (4)
& (5)
& 
& 
& \\[2mm]

\myrowcolour
\textbf{FireNet}     
& Precision 
& 65.22\%
& 72.53\%
& 28.33\%
& \textbf{78.60\%}
& 69.24\%
& Total Set
& 502
&\\[1mm]
\myrowcolour
& F1-Score 
& 16.33\%
& 45.88\%
& 06.96\%
& \textbf{83.44\%}
& 40.99\%
& Training Set
& 350
&\\[1mm]
\myrowcolour
& Recall 
& 09.11\%
& 33.56\%
& 03.96\% 
& \textbf{62.74\%}
& 31.74\%
& Validation Set
& 100
&\\[1mm]
\myrowcolour
& mAP(50)  
& 39.56\%
& 53.44\%
& 15.74\%
& \textbf{31.59\%}
& 49.37\%
& Test Set
& 50
& \\[1mm]
\myrowcolour
& mAP(50-95)
& 17.34\%
& 27.00\%
& 08.08\%
& \textbf{24.75\%}
& 25.50\%
& Out of Set
& 2
& \\[1mm]
\myrowcolour
& Rank
& (4)
& (1)
& (5)
& \textbf{(3)}
& (2)
& 
& 
& \\[2mm]

\textbf{Flame3}     
& Precision
& 67.41\%
& 70.25\%
& 58.89\%
& 81.11\%
& \textbf{96.40\%}
& Total Set
& 738
\\[1mm]
& F1-Score 
& 50.88\%
& 41.23\%
& 56.21\%
& 79.09\%
& \textbf{88.25\%}
& Training Set
& 500
&\\[1mm]
& Recall 
& 43.94\%
& 39.77\%
& 53.33\%
& 77.17\%
& \textbf{85.60\%}
& Validation Set
& 145
&\\[1mm]
& mAP(50)   
& 54.23\%
& 51.19\%
& 51.22\%
& 75.29\%
& \textbf{76.91\%}
& Test Set
& 90
& \\[1mm]
& mAP(50-95)
& 40.55\%
& 42.22\%
& 44.28\%
& 64.44\%
& \textbf{68.33\%}
& Out of Set
& 3
& \\[1mm]
& Rank
& (3)
& (5)
& (4)
& (2)
& \textbf{(1)}
& 
& 
& \\[2mm]

\bottomrule
 \end{tabular}}
\end{table}





\section{Conclusion }
\label{sec: Conclusion}
Over the past two decades, substantial progress has been made in fire threat management, encompassing prevention, mitigation, assessment, and control, driven by advancements in AI and computer vision technologies. This paper provides an extensive review of 36 fire and smoke datasets, analyzing their characteristics, collection methods, and imaging modalities to assess their suitability for various fire management applications. Our study highlights these datasets' key strengths and limitations, offering insights into their effectiveness in supporting DL-based detection, segmentation, and classification tasks. Additionally, we conducted comprehensive experimental evaluations using some powerful and widely used state-of-the-art models to benchmark dataset performance. These evaluations revealed the key potentials of each dataset, as well as the existing gaps and challenges that affect the generalizability of the model across diverse fire scenarios. By systematically examining these datasets, this study contributes to the ongoing development of more robust and adaptable fire detection and monitoring systems, paving the way for future improvements in data-driven fire and smoke management solutions.

Despite the significant progress in fire and smoke dataset collection, several key challenges still remain unaddressed, which would be valuable opportunities for future research and direction. One critical direction is enhancing dataset diversity by incorporating more geographically varied and multi-seasonal data to improve model generalization across different environmental conditions. Furthermore, integrating spatio-temporal information from multi-modal sensors, such as LiDAR and satellite imagery, can provide deeper insights into fire behavior dynamics. Future efforts should also prioritize developing standardized benchmarks and evaluation protocols to ensure consistent comparisons across various datasets and models. Last but not least, domain adaptation techniques and synthetic data generation approaches could help to solve data shortages and improve model robustness in real-world scenarios. Addressing these challenges will be crucial in advancing AI-driven fire and smoke management systems, ensuring their efficiency and adaptability to the dynamic challenges of wildfire management.

\section*{Declaration of Competing Interest}
The authors declare that they have no known competing financial interests or personal relationships that could have appeared to influence the work reported in this paper.

\section*{Acknowledgement}
\label{sec:Acknowledgement}
This material is based upon work supported by the National Aeronautics and Space Administration (NASA) under award number 80NSSC23K1393, and the National Science Foundation under Grant Numbers CNS-2232048, CNS 2120485 and CNS-2204445. 

\bibliographystyle{elsarticle-num} 
\bibliography{references}
\end{document}